\documentclass[sn-mathphys,Numbered]{sn-jnl}

\usepackage{graphicx}%
\usepackage{multirow}%
\usepackage{amsmath,amssymb,amsfonts}%
\usepackage{enumerate}
\usepackage{amsthm}%
\usepackage{mathrsfs}%
\usepackage[title]{appendix}%
\usepackage{xcolor}%
\usepackage{textcomp}%
\usepackage{manyfoot}%
\usepackage{booktabs}%
\usepackage{algorithm}

\usepackage{algorithmicx}%
\usepackage{algpseudocode}%
\usepackage{listings}%
\usepackage{natbib}
\usepackage{fancyvrb,cprotect}

\theoremstyle{thmstyleone}%
\theoremstyle{thmstyletwo}%

\theoremstyle{thmstylethree}%

\begin{document}

\title[Article Title]{Occluded Human Pose Estimation based on Limb Joint Augmentation}


\author[1]{\fnm{Gangtao} \sur{Han}}

\author[2]{\fnm{Chunxiao} \sur{Song}}

\author*[1,3]{\fnm{Song} \sur{Wang}}\email{ieswang@zzu.edu.cn}

\author[1]{\fnm{Hao} \sur{Wang}}

\author[1]{\fnm{Enqing} \sur{Chen}}

\author[3]{\fnm{Guanghui} \sur{Wang}}

\affil[1] {\orgdiv{School of Electrical and Information Engineering}, \orgname{Zhengzhou University}, \orgaddress{\street{100 Kexue Avenue}, \city{Zhengzhou}, \postcode{450001}, \state{Henan}, \country{China}}}
\affil[2] {\orgdiv{School of Computer and Artificial Intelligence}, \orgname{Zhengzhou University}, \orgaddress{\street{100 Kexue Avenue}, \city{Zhengzhou}, \postcode{450001}, \state{Henan}, \country{China}}}
\affil[3] {\orgdiv{Department of Computer Science}, \orgname{Toronto Metropolitan University}, \orgaddress{\street{350 Victoria Street}, \city{Toronto}, \postcode{M5B 2K3}, \state{ON}, \country{Canada}}}


\abstract{Human pose estimation aims at locating the specific joints of humans from the images or videos. While existing deep learning-based methods have achieved high positioning accuracy, they often struggle with generalization in occlusion scenarios. In this paper, we propose an occluded human pose estimation framework based on limb joint augmentation to enhance the generalization ability of the pose estimation model on the occluded human bodies. Specifically, the occlusion blocks are at first employed to randomly cover the limb joints of the human bodies from the training images, imitating the scene where the objects or other people partially occlude the human body. Trained by the augmented samples, the pose estimation model is encouraged to accurately locate the occluded keypoints based on the visible ones. To further enhance the localization ability of the model, this paper constructs a dynamic structure loss function based on limb graphs to explore the distribution of occluded joints by evaluating the dependence between adjacent joints. Extensive experimental evaluations on two occluded datasets, OCHuman and CrowdPose, demonstrate significant performance improvements without additional computation cost during inference.
}



\maketitle
\newpage
\section{Introduction}\label{sec1}

Human pose estimation, as a long-standing problem in computer vision, involves locating the human body joints in images or videos. It has attracted board interest due to its benefits for many applications, such as action recognition \cite{bib3}, character interaction \cite{bib2}, and human behavior generation \cite{bib1}. In recent years, deep learning-based techniques have witnessed remarkable progress in human pose estimation \cite{newell2016stacked,chu2017multi, bib14,bib15,xu2024disentangled}. However, most state-of-the-art methods are designed for scenarios with sparse crowds, where only a few people are distributed across input images, and they often experience significant performance degradation when faced with occluded scenarios \cite{khirodkar2021multi}.

In the real world, it is common for parts of the human body to be occluded by objects or other people when captured by cameras, posing a significant challenge for existing pose estimation methods. Addressing this challenge directly and intuitively involves collecting a large number of occluded samples with annotations to train pose estimation models. However, annotating occluded samples is an expensive and time-consuming process. To overcome the scarcity of annotated occluded samples, some researchers have turned to Generative Adversarial Networks (GANs) to generate plausible pose configurations and infer the poses of occluded body parts. For instance, Peng et al. \cite{peng2018jointly} utilize a generator to synthesize challenging samples and a discriminator as the pose estimation network to predict keypoints from these challenging samples. While GAN-based methods have improved estimation performance in occlusion scenarios, they often face challenges related to unstable and time-consuming training. To mitigate these issues, recent research explores the context of the image and the human body's structure based on visible joints. Iqbal and Gall \cite{iqbal2016multi} frame occluded human pose estimation as a joint-to-person association problem, capturing human body poses even in the presence of severe occlusions. Chen et al. \cite{chen2018cascaded} propose a Cascade Pyramid Network consisting of a GlobalNet to localize visible keypoints and a RefineNet to predict invisible joints by integrating multi-scale features from the GlobalNet. 

Building upon multi-scale features, Su et al. \cite{su2019multi} embed the attention mechanism into the network to emphasize useful context information in the integrated features. Recognizing the powerful ability of transformers to perceive global dependencies among keypoints, Li et al. \cite{li2021tokenpose} introduce a pure transformer-based model, namely TokenPose, to capture visual appearance relations and constraint cues from interactions between keypoint tokens and visual tokens. To reduce computation complexity, Ma et al. \cite{ma2022ppt} propose the Token-Pruned Pose Transformer for pruning background tokens and locating the human body area. Departing from traditional two-stage methods, PETR \cite{shi2022end} establishes an end-to-end multi-person pose estimation framework with transformers, predicting instance-aware full-body poses directly. While these methods have achieved significant progress in occluded human pose estimation, the advancements often come at the cost of designing intricate and complex network architectures, imposing a heavy computational burden during model inference.

To alleviate the computation burden, this paper introduces an occluded human pose estimation framework based on limb joint augmentation, aiming to enhance the performance of pose estimation models in occlusion scenarios. The proposed framework involves the design of occlusion blocks to conceal limb joints in training images, as illustrated in Fig. \ref{fig1}. These obscured images are treated as augmented samples for training the human pose estimation model. Leveraging prior knowledge provided by the human body structure, the paper proposes a Dynamic Structure Loss (DSL) function to emphasize the dependence between adjacent keypoints in human limbs. Consequently, the proposed framework enhances the performance of pose estimation models on occluded humans without introducing additional computational costs during inference. 

The primary focus of this paper is to investigate the impact of limb joints on occluded pose estimation. This emphasis arises from two main reasons. Firstly, limb joints are the most commonly occluded target in the pose estimation datasets. Of course, there are occluded heads in the collected images. However, the human bodies are hardly annotated and detected when the heads are occluded. There is no pose estimation task if the human bodies are not detected. Secondly, the predicted positions for limb joints are prone to have large shifts to their real positions. Since the head is a rigid structure, the limb joints have a larger degree of freedom than the head joints, resulting in a wider distribution range for the limb joints. It is difficult to accurately locate the occluded limb joints in a wide area, leading to performance degradation in pose estimation.

The contributions of this paper are summarized as follows:
\begin{enumerate}[(1)]
    \item The paper introduces an occluded human pose estimation framework based on limb joint augmentation to enhance the performance of pose estimation models in occlusion scenarios. The proposed framework achieves this without incurring additional computation costs during inference.

    \item The limb joint augmentation strategy is devised to simulate scenarios where objects or other people partially occlude the human body. Through the augmentation of training samples, the pose estimation model is incentivized to accurately determine the positions of occluded joints based on the information from visible ones.
    
    \item A limb structure loss function is constructed based on limb graphs, exploring the dependence between adjacent joints on human limbs. To stabilize the optimization process, a dynamic weighting scheme is designed to gradually enhance the constraint of the limb structure during the training phase. 
    
    \item To demonstrate the effectiveness of the proposed method, this paper conducts extensive experiments on two occluded datasets, namely OCHuman and CrowdPose. The experimental results illustrate that the proposed method achieves significant performance improvements.
\end{enumerate}

The rest of this paper is organized as follows. Section \ref{sec2} reviews the related work. The proposed method is detailed in Section \ref{sec3}. Section \ref{sec4} evaluates the performance of the proposed framework on occluded datasets, followed by the conclusion in Section \ref{sec5}.

\section{Related Work}\label{sec2}

\subsection{Human Pose Estimation}\label{subsec2.1}
Existing deep learning-based methods can be roughly divided into top-down and bottom-up two categories \cite{zheng2023deep}.  
The top-down methods contain two steps to estimate the positions of the human body joints. At first, a person detector is employed to obtain a set of body boxes from the input images, where each box corresponds to one person. A pose estimator then detects the joints for each person's box. Current top-down research focuses on developing effective pose estimator architectures and exploring more supervision information from the input images. Newell et al. \cite{newell2016stacked} build the stacked hourglass network to integrate multi-scale feature maps. Following \cite{newell2016stacked}, hourglass residual units are designed in \cite{chu2017multi} to expand the receptive fields and capture features from various scales. Wei et al. \cite{wei2016convolutional} design convolutional pose machines to sequentially refine the joint predictions from the 2D belief maps. Sun et al. \cite{bib8} propose to connect multi-resolution subnetworks in parallel and conduct repeated multi-scale fusions to learn high-resolution representations. Based on the multi-resolution parallel design in \cite{bib8}, HRFormer \cite{bib13} proposes a high-resolution converter to extract features. Yang \cite{bib14} et al. utilize the attention layer as an aggregator to capture the position dependence of the joints. Besides these efforts in designing network architectures, existing methods also propose to explore more supervised information. Chu et al. \cite{chu2016structured} introduce a structured feature-level learning framework to reason the correlations among human body joints. A structure-aware loss function is designed in \cite{ke2018multi} to improve the matching of keypoints and respective neighbors. Tang et al. \cite{tang2018deeply} build a deeply learned compositional network to model the realistic relationship among body parts. 

Different from the top-down methods, the bottom-up ones locate all the visible joints in the input images first. After that, the located joints are grouped into individual persons. DeepCut \cite{pishchulin2016deepcut} is a classical bottom-up approach that utilizes the Fast R-CNN-based body part detector to capture all the body part candidates and the integer linear programming to label and assemble these parts. In OpenPose \cite{cao2017realtime},  convolutional pose machines \cite{wei2016convolutional} are employed to predict keypoint coordinates. To associate the keypoints to each person, OpenPose further designs Part Affinity Fields to encode the position and orientation of limbs. Cheng et al. \cite{bib7} extend HRNet \cite{bib8} and propose to deconvolve high-resolution heatmaps to extract body features. LitePose \cite{bib9} introduces a converged deconvolution head to eliminate redundant refinement in high-resolution branches. To reduce quantization errors in post-processing, Wang et al. \cite{bib10} utilize super-resolution lightweight heads to predict high-resolution heatmaps. Typically, the bottom-up methods enjoy low computation complexity but suffer from poor performance in occluded scenarios. Considering the requirement for positioning accuracy in practical applications, top-down methods are more popular than their rivals. Apart from the above two categories, some end-to-end pose estimation frameworks have been proposed in recent years. For example, PETR \cite{shi2022end} views pose estimation as a hierarchical set prediction problem and removes the hand-crafted modules.

\subsection{ Occluded Human Pose Estimation} \label{subsec2.2}

Although most existing human pose estimation models have achieved promising performance, they still face the challenge of occlusion. Some strategies are proposed to address these challenges. For instance, several researchers propose to create occluded datasets to train the pose estimation models. Zhang et al. \cite{bib35} introduce a ``OCcluded Human'' (OCHuman) dataset to provide heavy occlusion samples for model training. Li et al. \cite{bib36} collect 20,000 images from different public datasets, and create the CrowdPose dataset with a uniform distribution of crowd index. With the help of occluded datasets like OCHuman or CrowdPose, the capability of current pose estimation models has been improved to handle the keypoint detection for occluded human bodies to some extent. However, the existing occluded datasets cannot cover all kinds of occlusion types in the real world. In addition, data collection and annotation are very expensive and time-consuming, hardly used in various real applications \cite{zhang2021six}. 

To reduce the cost of obtaining occluded samples, GAN-based methods are developed to generate synthesized data \cite{peng2018jointly,li2020cascaded,xu2021domain}. An alternative approach is to expand the receptive field of the networks, simulating the way that human beings deduce the positions of the occluded keypoints from images. Chen \cite{chen2018cascaded} and Su \cite{su2019multi} aggregate multi-scale features to perceive the global context information from the input images. Taking advantage of the capability of the Transformer to perceive global dependencies of the keypoints, TokenPose \cite{li2021tokenpose} is proposed to capture global visual appearance relations in the first few layers, and gradually converge to local regions as the network goes deeper. Token-Pruned Pose Transformer \cite{ma2022ppt} prunes background tokens to make the network focus on the human body area. Although these aforementioned methods have made great progress in occluded human pose estimation, they suffer from heavy computation complexity, leading to a limited application in industry. 

In this paper, we propose a simple but effective occluded human pose estimation framework to improve the occluded joint positioning accuracy without introducing extra computational costs for model inference.

\section{The Proposed Framework}\label{sec3}
This section describes the details of the proposed framework for occluded human pose estimation. We first present the overall framework in Sec. \ref{subsec3.1}, followed by the limb joint augmentation strategy in Sec. \ref{subsec3.2}, and the dynamic structure loss in Sec. \ref{subsec3.3}.

\subsection{Overall Framework}\label{subsec3.1}
\begin{figure}[htbp]
	\centering{\includegraphics[width=1.0\textwidth]{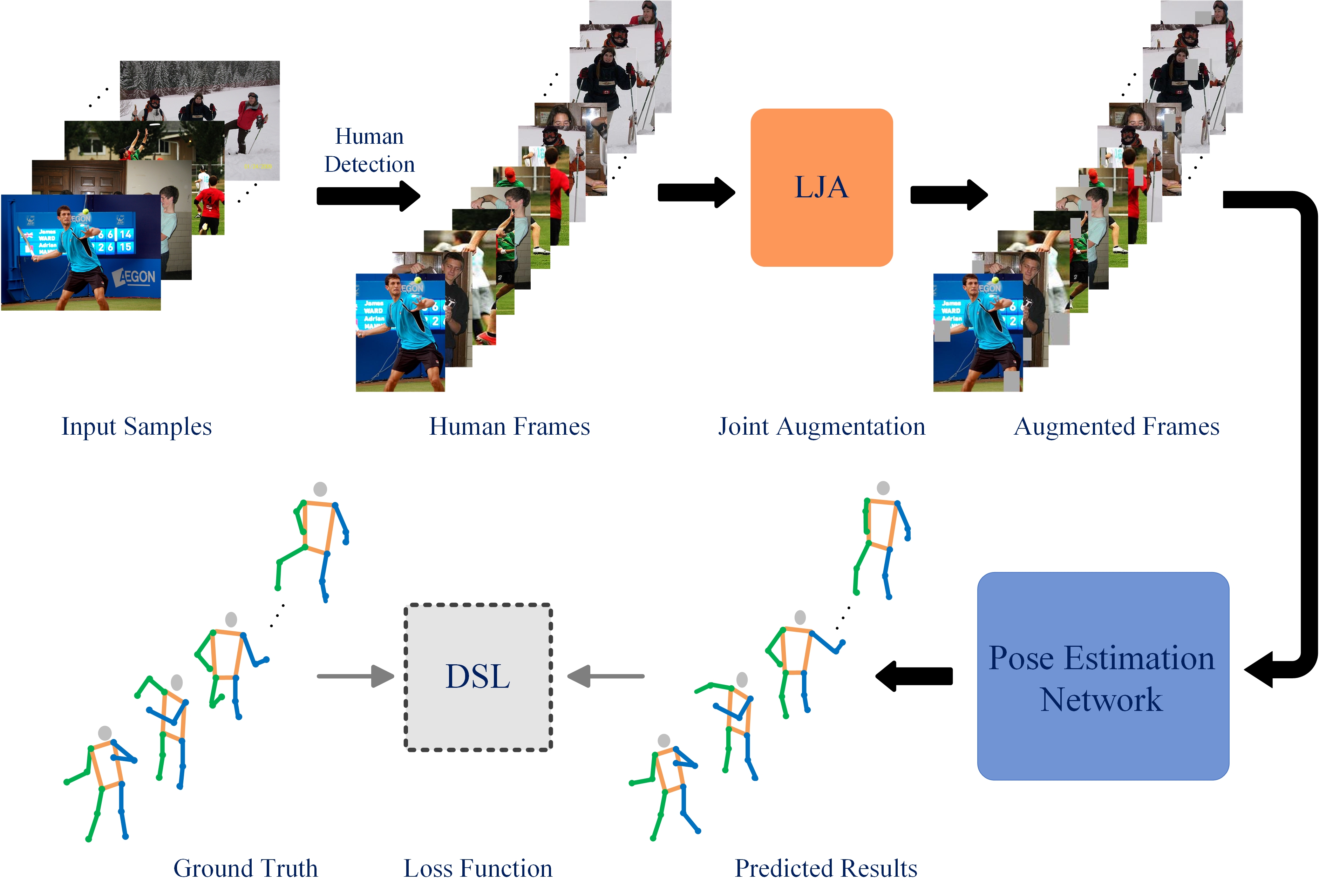}}
	\caption{The overall framework. This paper proposes to simulate the real occlusion scene by generating occlusion blocks over the limb joints. The pose estimate network takes the occluded human frames as input and predicts the positions of all the joints. The dynamic structure loss explores the dependence between adjacent joints on limbs. With the augmented training samples and the constraint of human structure, the network generates accurate prediction results for the occluded joints.\label{fig1}}
\end{figure}

The overall framework of the proposed method is illustrated in Fig. \ref{fig1}. Following the top-down methods, this paper first acquires the human frames from the input images. For the human frames, this paper proposes to augment the visible limb joints using the occlusion blocks. The occlusion blocks are generated based on the joints' locations and the sizes of the human frames. After that, the augmented human frames with occluded joints are fed into the pose estimation network. Based on the prior knowledge of human body structures, the dynamic structure loss forces the network to accurately predict the positions of the occluded joints by perceiving the dependence of the adjacent joints on limbs.

\subsection{Limb Joint Augmentation }\label{subsec3.2}

In the real world, it is inevitable that the human bodies are partially occluded by other ones or objects when shot by cameras. Existing pose estimation models trained by the single-person images hardly achieve satisfactory performance in the occlusion scenario. To improve the joint detection performance, this paper proposes a limb joint augmentation strategy. Unlike traditional augmentation methods which transform the input images in geometry to increase the number of the training samples, the proposed limb joint augmentation strategy aims at generating joint-occlusion training samples. 

According to the joint annotations, human joints are divided into visible and occluded ones. The proposed method mainly focuses on the visible limb joints. For each image, the visible limb joints are formulated as a set $\mathcal{P}=\left\{P_1,P_2, \cdots ,P_V \right\}$, where ${V}$ represents the number of visible joints. To augment the image, we randomly select ${v}$ visible limb joints to form the occlusion joint set  $\mathcal{Q}=\left\{Q_1,Q_2, \cdots,Q_v \right\}$. Since the numbers of visible limb joints $V$ in different images are various, we set the value of $v$ as a fixed ratio of $V$
 \begin{align}
      v=\lceil\alpha \cdot V \rceil\label{eq1},
\end{align}
where ${\alpha}$ is the fixed ratio. The symbol ${\lceil \  \rceil} $ represents that the value of $\alpha \cdot V$ is rounded up.
For each selected limb joint, its corresponding occlusion block is generated with the joint's coordinate as the center. The size of the occlusion block is set as a downscale of the human frame. Specifically, the height $h_o$ and width $w_o$ of the occlusion block are defined as follows
\begin{align}
	\left[h_o,w_o\right] =\beta \cdot \left[h,w\right]\label{eq2},
\end{align}
where ${\beta}$ represents the downscaling ratio, and ${h}$ and ${w}$ denote the height and width of the body frame respectively. Thus, the occlusion area ${R}$ at the joint is obtained as follows
\begin{align}
	R=(x-w_o/2, y-h_o/2, x+w_o/2, y+h_o/2)\label{eq3},
\end{align}
where $(x,y)$ represents the coordinate of occluded joints. The values of the occlusion block remain the same, which is assigned randomly from the range of $[0, 255]$.  

\begin{figure}[!t]
	\centering{\includegraphics{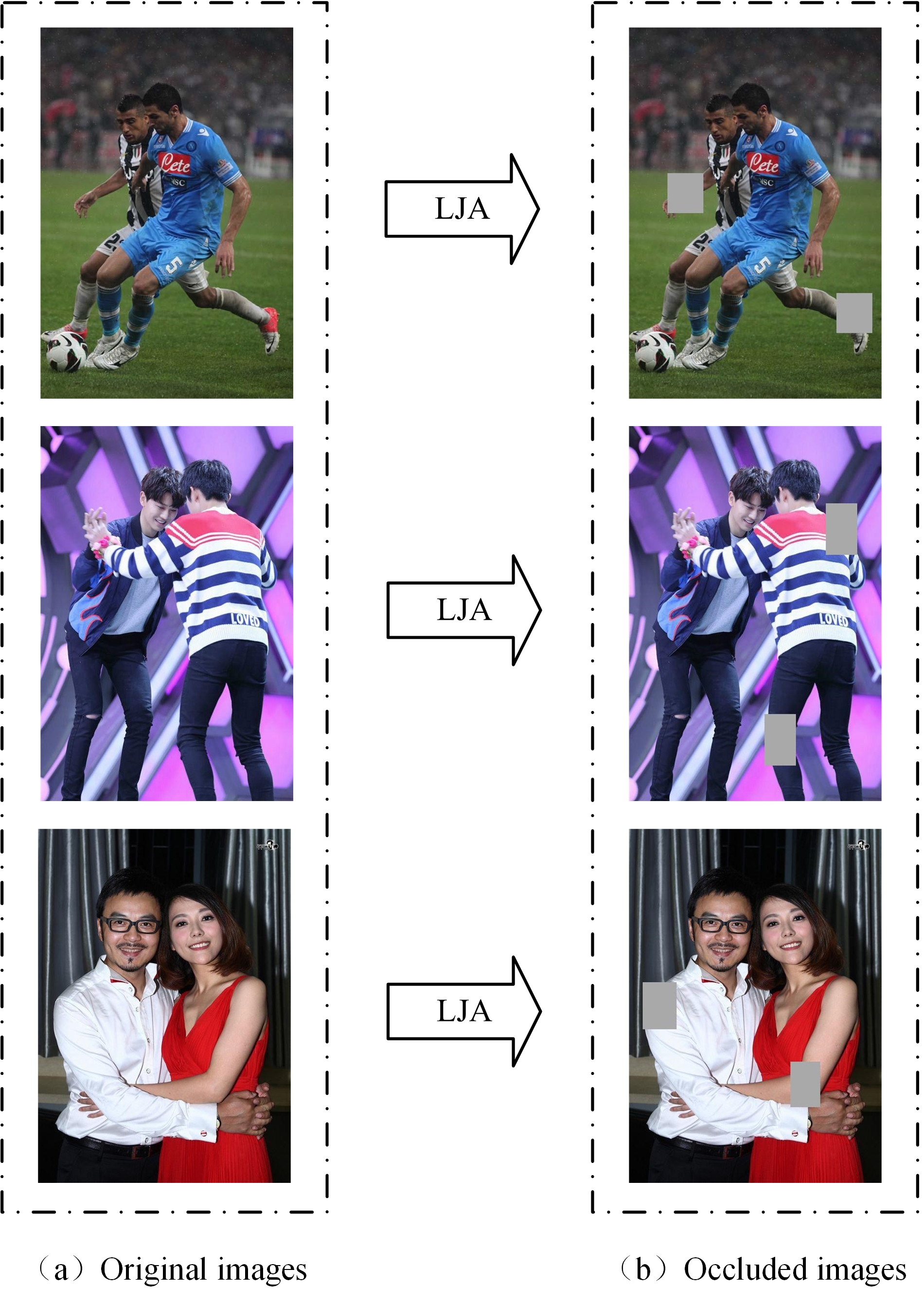}}
	\caption{The visualization of limb joint augmentation (LJA). (a) Original images. (b) Occluded images generated by LJA. In the experiments, we set ${\alpha=0.15}$, and ${\beta=0.20}$. The values of the occlusion area in the above images are set to 169 for clear visualization. \label{fig2}}
\end{figure}

The results of Limb Joint Augmentation (LJA) on whole images are illustrated in Fig. 2. For clearer visualization, we set the values of the occluded areas to 169 in Fig. 2. In our experiments, however, the values of the occluded areas are randomly assigned within the range of [0, 255] to avoid any artifacts or biases that might result from using a fixed value. By augmenting the visible human joints, LJA avoids the need for exhaustive annotation of occluded images and the use of complex network designs to enhance pose estimation performance. It simulates real scenarios where human bodies are partially occluded and generates a set of plausible occluded samples for model training. These generated samples help train the pose estimation model to be more robust in handling occluded human bodies.

\subsection{Dynamic Structural Loss}\label{subsec3.3}
Pose estimation techniques can be broadly categorized into regression methods and heatmap-based methods based on the types of prediction results. Regression methods employ end-to-end networks to directly predict joint coordinates from images. The goal of heatmap-based methods is to predict the 2D heatmaps which are required to be the same as the ground-truth heatmaps. The ground-truth heatmaps are generated by filtering each real joint's location with 2D Gaussian kernels \cite{bib33}. Compared with regression methods, the training of the heatmap-based methods is smoother because the generated heatmaps preserve the spatial location information \cite{zheng2023deep}. Building upon the heatmap-based methods, this paper proposes a dynamic structure loss which is defined as
\begin{align}
	L_{DSL}=L_{MSE}+\lambda L_{LSL} \label{eq4},
\end{align}
where $L_{MSE}$ denotes the standard heatmap-based loss. The limb structure loss $L_{LSL}$ focuses on exploring the dependency of adjacent joints in limbs. Throughout the training process, the dynamic weighting factor $\lambda$ serves to balance the influences of $L_{MSE}$ and $L_{LSL}$ on the model training

\begin{figure}[htbp]
	\centering{\includegraphics[scale=0.9]{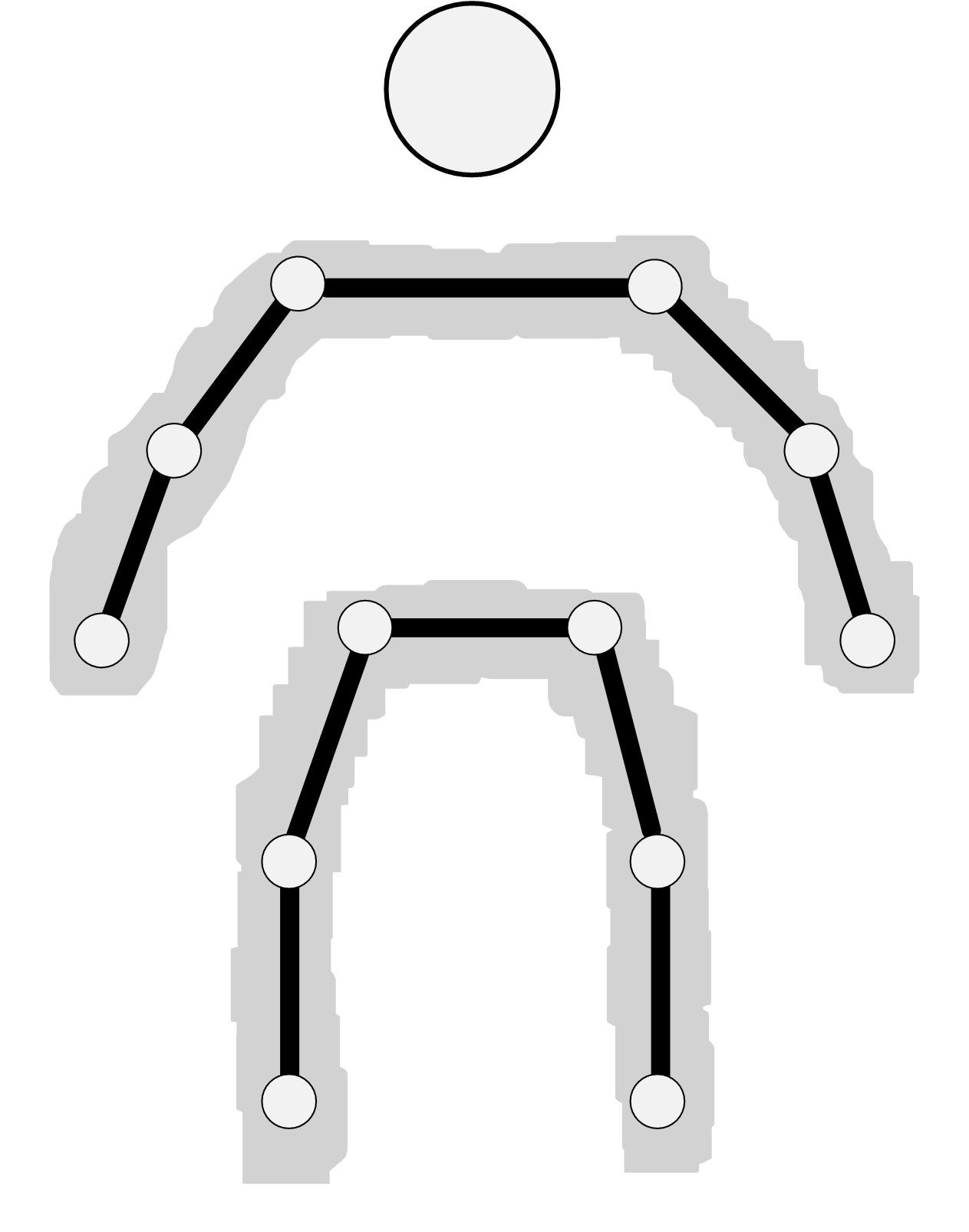}}
	\caption{The visualization of human limb graphs with joints. This paper constructs two separate limb graphs: \textit{left wrist-left elbow-left shoulder-right shoulder-right elbow-right wrist}, and \textit{left ankle-left knee-left hip-right hip-right knee-right ankle}.
 \label{fig3}}
\end{figure}

The standard heatmap-based loss evaluates the Mean Square Error (MSE) between the predicted heatmaps and the ground-truth ones. 
\begin{align}
	L_{MSE}=\frac{1}{N}\sum_ {i=1}^N \lVert P_i-G_i\rVert^2   \label{eq5},
\end{align}
where ${P_i}$ and ${G_i}$ represent the predicted and ground-truth heatmaps of the $i$-th joints respectively. ${N}$ denotes the number of joints. From Eq.\eqref{eq5}, it is known that the standard heatmap-based loss focuses on optimizing the predicted positions of individual joints.

\begin{figure}[htbp]
	\centering{\includegraphics[width=0.8\textwidth]{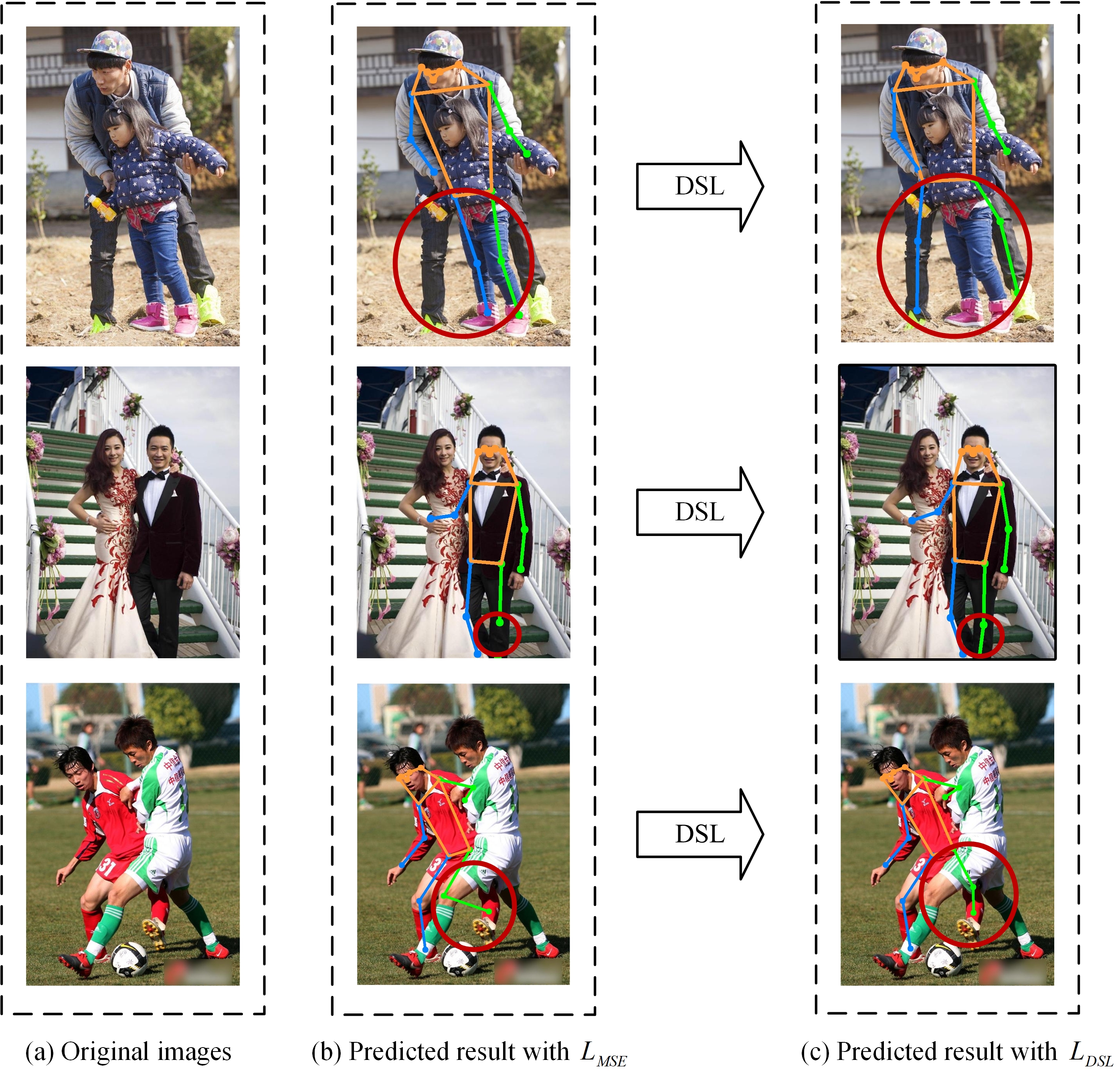}}
	\caption{Visualization of pose estimation results after correction by dynamic limb structure loss. \label{fig4}}
\end{figure}

Inspired by \cite{ke2018multi}, this paper constructs a limb structure loss $L_{LSL}$ to locate the occluded joints by exploring the dependency of adjacent limb joints. Based on the observation of the human's daily actions, it is well known that the joints on the same limb usually have a high correlation with each other. Thus, this paper constructs two separate limb graphs: \textit{left wrist-left elbow-left shoulder-right shoulder-right elbow-right wrist}, and \textit{left ankle-left knee-left hip-right hip-right knee-right ankle}. The constructed human joint graphs are illustrated in Fig. \ref{fig3}. It is evident that every joint on the limbs is connected by one or two other joints. For example, the left shoulder is related to the right shoulder and the left elbow. The right foot is only connected with the right knee. Based on these observations, the proposed limb structure loss is defined as
\begin{align}
	L_{LSL}=\sum_{i=1}^M \lVert P\textquotesingle_i-G\textquotesingle_i\rVert^2  \label{eq6},
\end{align}
where ${P\textquotesingle_i}$ represents the predicted structure heatmap which is obtained by summing the heatmaps of the $i$-th keypoint and its neighbors. Similarly, the real structure heatmap ${G\textquotesingle_i}$ is calculated by adding up the real heatmap of the $i$-th joints and their neighbors. The parameter ${M}$ in Eq.\eqref{eq6} represents the number of joints in limbs. In our setting, the value of ${M}$ is set to 12. By minimizing the limb structure loss $L_{LSL}$, the position shifts of the joints are corrected. Fig. \ref{fig4} shows some examples of the predicted results by the dynamic human limb structure loss Eq.\eqref{eq4}.

\begin{figure}[htbp]
    \centering{\includegraphics[scale=0.6]{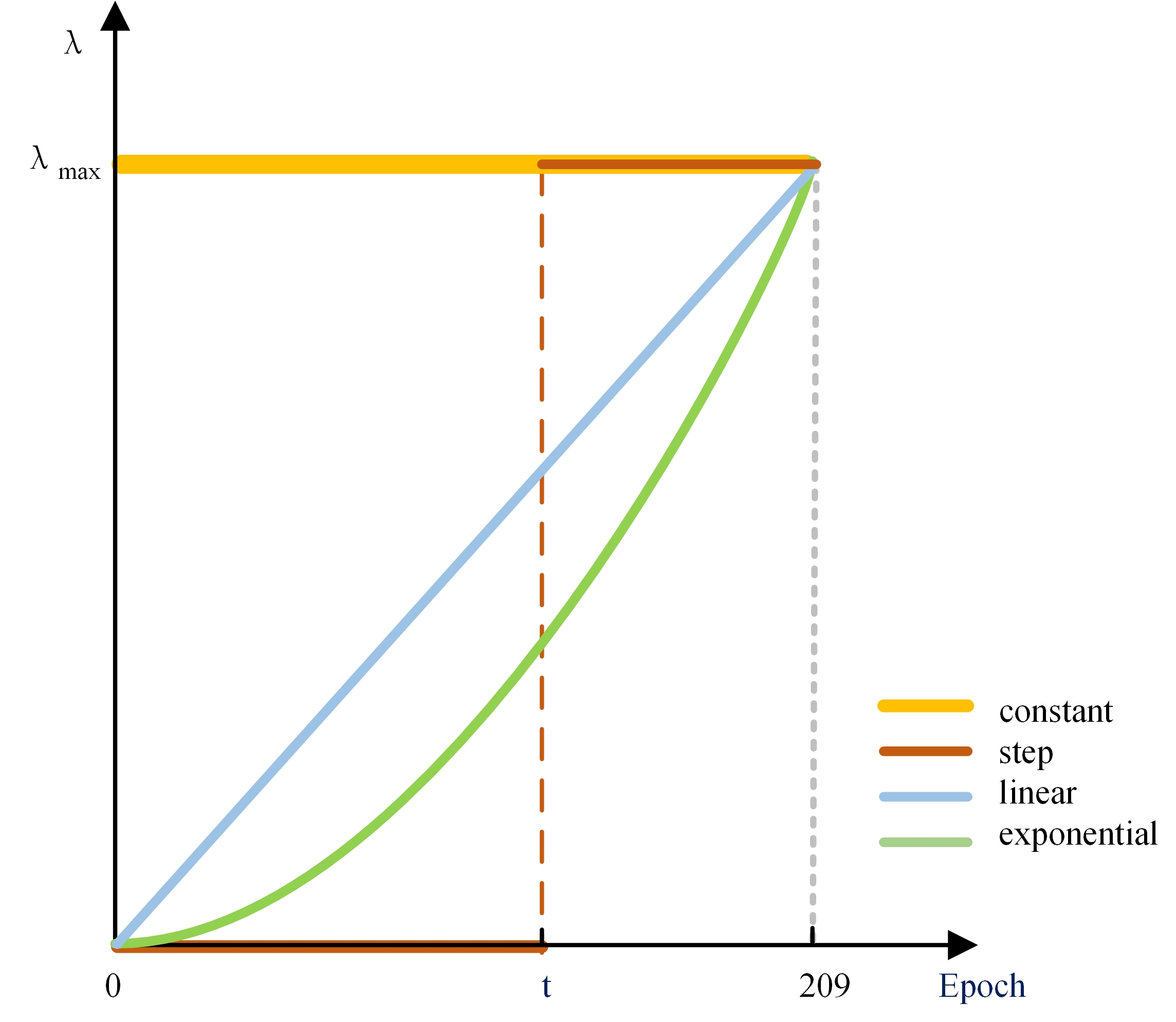}}
	\caption{The options of the weighting scheme, where ${\lambda}$ changes along with the increasing epoch.\label{fig5}}
\end{figure}

Considering that the network is not able to accurately localize the joints during the initial phase of training, this paper suggests gradually increasing the weighting factor $\lambda$. This adjustment is made because an excessively large $\lambda$ could lead to the network optimization oscillating between the individual joint's position and the limb structure, resulting in slow convergence. To circumvent this issue, the paper proposes dynamically weighting the limb structure loss $L_{LSL}$. Various weighting schemes, such as constant, step, linear, and exponential, are illustrated in Fig. \ref{fig5}. The impact of different weighting schemes on performance is discussed in Sec. \ref{subsec4.5}.

To summarize, this paper introduces a limb structure loss $L_{LSL}$ into the loss function to locate the occluded limb joints by leveraging prior knowledge of human body structures. Compared with the existing structure-aware loss in \cite{ke2018multi}, the proposed dynamic structural loss considers the possible large shifts of the occluded joints in limbs, and constructs limb graphs to constrain the shift range of the occluded joints. In addition, to stabilize the optimization process, a dynamic weighting scheme is designed to gradually enhance the constraint of the limb structure during the training phase.

\section{Experiments}\label{sec4}
\subsection{Datasets}\label{subsec4.1}
The performance of the proposed Limb Joint Occlusion Framework (LJOF) is evaluated on two occlusion datasets, i.e. OCHuman \cite{bib35}, and CrowdPose \cite{bib36}. 

\textbf{OCHuman dataset.} OCHuman dataset is a recently proposed benchmark for heavily occluded humans. It contains 4,731 annotated images with 17 keypoints, of which 2,500 images are for validation, and 2,231 images for testing. Following \cite{bib35}, we choose the COCO training set \cite{bib34} to train the models and report the performance on the OCHuman testing set. COCO dataset \cite{bib34} is a popular large benchmark for keypoint detection, including 118,287 images for training and 5,000 images for validation. Due to the large number of samples, the COCO dataset is typically used to train the pose estimation models. 

\textbf{CrowdPose dataset.} CrowdPose dataset contains 20,000 crowded images from other public datasets like COCO \cite{bib34}. It is divided into 10,000, 2,000, and 8,000 images for training, validation, and testing respectively. Following \cite{bib7,bib17}, we train the models on the training set and validation set, and evaluate them on the testing set.

\textbf{Evaluation metrics.} This paper follows the standard evaluation metrics and uses metrics based on OKS (object keypoint similarity) to evaluate the performance of the models on different datasets. We report average precision (AP), and average recall (AR) with different thresholds $AP^{50}, AP^{75}, AR^{50}, AR^{75}$ and object sizes $AP^{M}, AP^{L}$. For the CrowdPose dataset, we also report the AP at different crowding levels: $AP^{Easy}, AP^{Medium}, AP^{Hard}$.

\subsection{Implementation Details}\label{subsec4.2}
In this paper, all the experiments are implemented with MMPose\footnote{https://github.com/open-mmlab/mmpose}. The training epoch is set to 210 for all datasets. As for the other training settings, we use the default configures in MMPose for all the implemented models. In the limb joint augmentation, the parameters ${\alpha}$ and ${\beta}$ are set to 0.15 and 0.20. The range of the dynamic weighting factor $\lambda$ in DSL is set to $[0,1 \times e^{-4}]$. The step weighting scheme is selected, where $\lambda$ steps to be $1\times e^{-4}$ when $Epoch=139$.

\subsection{Experiments on OCHuman}
\textbf{Comparison with baselines.} This paper plugs the state-of-the-art heatmap-based pose estimation networks into the proposed Limb Joint Occlusion Framework (LJOF)

\begin{table}[htbp]
\footnotesize
	\caption{Comparison with the baseline methods on the OCHuman testing set with ground truth bounding boxes.}\label{tab1}
	\begin{tabular}{@{}lllllllll@{}}
		\toprule
		Methods & Backbone &  Input Size& $AP$ & $AP^{50}$ & $AP^{75}$ & $AR$  &$ AR^{50}$&$AR^{75}$ \\
		\midrule	
		SimpleBaseline  \cite{bib11}      & ResNet-152 & 256x192 & 57.0 & 72.5 & 61.7 & 61.6 & 75.4 & 65.8    \\
		SimpleBaseline+LJOF                           & ResNet-152 &256x192 & {\bfseries 58.6} &	{\bfseries74.3}& {\bfseries63.7}& {\bfseries63.1}&	{\bfseries77.7}&	 {\bfseries 67.8} 	\\
		\midrule           		
		HRNet \cite{bib8}            & HRNet-W32  &256x192  & 59.1 & 74.8 & 64.1 & 63.1 & 77.5 & 67.6    \\	
		HRNet+LJOF                             & HRNet-W32  &	256x192	& {\bfseries60.6} & {\bfseries75.9} & {\bfseries66.2} & {\bfseries64.8} &  {\bfseries78.8}&{\bfseries69.7}\\ 
		\midrule 		
		HRNet  \cite{bib8}              & HRNet-W48  &	256x192	& 61.1 & 75.2 &	66.3 & 64.8 & 77.8 & 69.0  \\
		HRNet+LJOF                             & HRNet-W48  & 256x192 & {\bfseries63.5} & {\bfseries 77.1} &	 {\bfseries 69.5} &	 {\bfseries 67.1} & {\bfseries 79.4} &	{\bfseries 72.0}\\		
		\botrule	
		ViTPose-B*          & ViT-B	  & 256x192 & 59.5 & 74.3 & 65.6 & 63.9 & 77.1 & 69.3    \\ 		
		ViTPose-B+LJOF                            & ViT-B      & 256x192 & {\bfseries61.0} & {\bfseries76.6} & {\bfseries66.8}& {\bfseries65.2} & {\bfseries79.0} & {\bfseries70.4}   \\
  \midrule 
        SRPose*               & HRNet-W32  & 256x192 &  61.4&	75.2&	  66.5	&65.3	&  77.9	& 70.0\\
        SRPose+LJOF           & HRNet-W32  & 256x192 &   {\bfseries 62.6 }& {\bfseries  76.9 } & {\bfseries 68.1   }& {\bfseries  66.6}& {\bfseries    79.7 }& {\bfseries   71.4 }\\
		
		\botrule
	\end{tabular}
\end{table}

\noindent to improve their occluded keypoint detection performance. The selected pose estimation networks includes SimpleBaseline \cite{bib11}, HRNet-W32 \cite{bib8}, HRNet-W48 \cite{bib8}, ViTPose-B \cite{bib16}, and SRPose \cite{wang2023lightweight}. Table \ref{tab1} presents the result of plugging the existing networks into the proposed framework. The annotated human boxes in OCHuman are utilized for evaluation. In Table \ref{tab1}, the symbol ``*" indicates that its results are obtained by re-implementing the official code. We utilize ``+LJOF'' to represent that the network is plugged into the proposed framework. From Table \ref{tab1}, we can see that the proposed framework achieves the best performance with a margin of $1.6\%$, $1.5\%$, $2.4\%$, $1.5\%$, $1.2\%$ in $AP$ over SimpleBaseline, HRNet-W32, HRNet-W48, ViTPose-B, and SRPose, respectively. 

\textbf{Comparison with SOTAs.} To verify the superiority of the proposed framework,  conducts comparisons with state-of-the-art (SOTA) methods, including HRFormer \cite{bib13}, PINet \cite{wang2021robust}, and BoIR \cite{jeong2023boir}. The comparison results are shown in Table \ref{tab_sota_oc}. The results clearly indicate that the proposed framework outperforms SOTAs with the same backbones. In comparison to HRFormer with the backbone ``HRFormer'', the proposed ``HRNet-W32+LJOF'' achieves better $AP$ performance with fewer parameters and Flops.

\begin{table}[ht]
\footnotesize
	\caption{Comparison with SOTAs on the OCHuman testing set with ground truth bounding boxes.}\label{tab_sota_oc}%
	\begin{tabular}{@{}llllllll@{}}
		\toprule
		Methods & Backbone &  Input Size& Params(M)     & FLOPs(G)& $AP$ &  $AP^{50}$ &  $AP^{75}$  \\
		\midrule
        HRFormer \cite{bib13}      & HRFormer-B & 256x192 &  43.2  & 12.2    & 49.8  &  70.4  &  53.4    \\
        HRFormer \cite{bib13}      & HRFormer-B & 384x288 &  43.2  & 26.8    & 49.7  &  71.6  &  52.4    \\
           \midrule
        PINet \cite{wang2021robust}           & HRNet-W32  & 512x512 &  29.4  & 42.4    & 59.8	&  74.9	 &  65.9   \\
        {\bfseries HRNet+LJOF}            & HRNet-W32  & 256x192 &28.5&7.1& {\bfseries60.6} & {\bfseries75.9} & {\bfseries66.2} \\	
        \midrule
        BoIR \cite{jeong2023boir}             & HRNet-W48  & 640x640 &  68.9  &  -      &48.5  &  61.3	 &  54.1    \\  
        {\bfseries HRNet+LJOF}            & HRNet-W48  & 256x192 &63.6&14.6& {\bfseries63.5} & {\bfseries 77.1} &	 {\bfseries 69.5} \\		
		\botrule
	\end{tabular}
\end{table}

\textbf{Evaluation under non-occluded scenario.} To validate that the proposed framework does not adversely affect the model's performance on non-occluded datasets, this paper further evaluates the trained models on the COCO validation set. The evaluation results are illustrated in Table \ref{tab2}. Similar to the experiments on OCHuman, this paper selects SimpleBaseline \cite{bib11}, HRNet-W32 \cite{bib8}, HRNet-W48 \cite{bib8}, ViTPose-B \cite{bib16}, and SRPose \cite{wang2023lightweight} as baselines. The official codes of ViTPose-B and SRPose are re-implemented for a fair comparison. Due to the limitation of hardware, we cannot train ViTPose-B and SRPose models with the same configuration as in \cite{bib16} and \cite{wang2023lightweight}. The symbol ``+LJOF'' is used to represent the proposed framework. From Table \ref{tab2}, it is observed that the proposed framework achieves slight performance gains on most of the metrics when compared with the baselines, demonstrating that the proposed framework has no detrimental impact on the performance of pose estimation models on non-occluded images.

\begin{table}[htb]
\footnotesize
	\caption{Comparison to the baseline methods on the COCO validation set.}\label{tab2}%
	\begin{tabular}{@{}lllllllll@{}}
		\toprule
		Methods                      & Backbone    &  Input Size & $AP$ & $AP^{50}$ & $AP^{75}$ & $AP^{M}$&$ AP^{L}$ &$AR$\\
		\midrule
		SimpleBaseline \cite{bib11}  & ResNet-152 &	256x192 & 72.0 & 89.3 &	79.8 & 68.7 & 78.9 & 77.8  \\
		SimpleBaseline+LJOF                         & ResNet-152 &256x192  &{\bfseries 73.7} & {\bfseries 90.6} &	{\bfseries 81.9} &	{\bfseries 70.0} & {\bfseries80.2} &{\bfseries  79.1}\\
		\midrule
  	HRNet  \cite{bib8}       & HRNet-W32  & 256x192 & 74.4 & {\bfseries 90.5}  &	{\bfseries 81.9} & 70.8 & {\bfseries 81.0} & 79.8 \\
		HRNet+LJOF                   & HRNet-W32  & 256x192& {\bfseries 74.5} &{\bfseries 90.5} & {\bfseries 81.9} & {\bfseries 71.0}  & {\bfseries 81.0} &	{\bfseries 79.9} \\
		\midrule
    	HRNet  \cite{bib8}      & HRNet-W48  & 256x192 & 75.1 & 90.6 &	82.2 & 71.5 & 81.8 & 80.4  \\
		HRNet+LJOF                        & HRNet-W48  & 256x192 & {\bfseries75.8 }& {\bfseries 90.8} & {\bfseries83.2} & {\bfseries 72.3} & {\bfseries 82.5} &  {\bfseries 81.1} \\
		\midrule
		ViTPose-B*      & ViT-B      &	256x192	& 75.7 & 90.5 &	83.0 & 72.1 & 82.4 & 80.9  \\
		ViTPose-B+LJOF                         & ViT-B      &	256x192	&{\bfseries 75.8} & {\bfseries 90.5} &	{\bfseries 83.2} & {\bfseries 72.4} & {\bfseries 82.5} & {\bfseries 81.1}  \\
    \midrule
       SRPose*                & HRNet-W32  & 256x192 &  75.8 &	90.6	&   82.4	&  72.3  &{\bfseries82.6	} & 80.9\\
       SRPose+LJOF            & HRNet-W32  & 256x192 &   {\bfseries 75.9 }   & {\bfseries90.8} & {\bfseries82.9 }&  {\bfseries72.5}&   82.4   &{\bfseries 81.1}\\
    
		\botrule
	\end{tabular}
\end{table}

\subsection{Experiments on CrowdPose}
\textbf{Comparison with baseline.} The evaluation results on CrowdPose are shown in Table \ref{tab3}. To validate the effectiveness of the proposed framework, this paper employs SimpleBaseline  \cite{bib11},  HRNet-W32 \cite{bib8}, HRNet-W48 \cite{bib8}, ViTPose-B \cite{bib16}, and SRPose \cite{wang2023lightweight} as the baselines. In particular, the models are trained on the CrowdPose training and validation sets. The experimental results are obtained by evaluating the models on the CrowdPose testing set. For a fair comparison, this paper retrains all the baselines. Due to the hardware limitation, we cannot set the same configuration of batch sizes as the default one. The images from CrowdPose contain more human instances than COCO when the same batch size is set. Therefore, the batch sizes are set to 32 for retraining the baseline models, while the default configuration is 64. Compared with the baselines, the proposed framework achieves gains of $0.5\%$, $3.2\%$, $2.9\%$ $0.6\%$, and $3.1\%$ in AP, demonstrating its generalization ability on occluded pose estimation. 

\begin{table}[htbp]
\footnotesize
	\caption{Comparison to the baseline methods on CrowdPose testing set. }\label{tab3}%
	\begin{tabular}{@{}lllllllll@{}}
		\toprule
		Methods & Backbone &  Input Size & $AP$ & $AP^{50}$ & $AP^{75}$&$AP^{E.}$  &$ AP^{M.}$&$AP^{H.}$ \\
		\midrule
        SimpleBaseline*                & ResNet-152       &256x192& 65.6 & 81.7	& 71.4	& 75.1  & 66.7 & 53.6\\
        SimpleBaseline+LJOF             & ResNet-152       &256x192& {\bfseries 66.1} & {\bfseries 82.0} & {\bfseries 72.2}  & {\bfseries 75.9}  & {\bfseries 67.2} & {\bfseries 54.1}\\
        \midrule
        HRNet*                          & HRNet-W32        &256x192& 64.2 &	81.0 & 69.5  & 74.1	 & 65.3	&51.8\\ 
        HRNet+LJOF                     & HRNet-W32        &256x192& {\bfseries 67.4} &	{\bfseries 82.4} & {\bfseries 72.8}  & {\bfseries 76.7}	 & {\bfseries 68.6} &{\bfseries 55.1}\\ 
        \midrule
        HRNet*                          & HRNet-W48        &256x192& 65.6 &	81.5 & 70.9  & 75.1  & 66.7 &53.4\\
        HRNet+LJOF                     & HRNet-W48        &256x192& {\bfseries 68.5} & {\bfseries 82.7} &{\bfseries 74.0}	 & {\bfseries 77.6}  & {\bfseries 69.7}	&{\bfseries 56.9}\\
            \midrule
		ViTPose-B*          & ViT-B	       &256x192& 66.5 & 81.5 & 72.3  & 76.1  & 67.9 & 54.6    \\ 	
		ViTPose-B+LJOF                             & ViT-B            &256x192&  {\bfseries67.1} &  {\bfseries82.8} & 	{\bfseries72.8} & 	{\bfseries76.7} & {\bfseries68.3} & {\bfseries55.3}\\
  \midrule
   SRPose*                        &  HRNet-W32 &  256x192   &64.7  & 81.1 & 69.8   &74.4 &  65.7 &  52.3 \\
        SRPose+LJOF                    &  HRNet-W32 &  256x192   & {\bfseries 67.8  } & {\bfseries82.5} & {\bfseries73.0 }&  {\bfseries77.0}&  {\bfseries 69.1}  &{\bfseries55.9}\\

		\botrule
	\end{tabular}
\end{table}

\textbf{Comparison with SOTAs.} We also compare the performance of the proposed framework with that of SOTAs on CrowdPose. The comparison results are presented in Table \ref{tab_sota_crowd}. It is evident from Table \ref{tab_sota_crowd} that the proposed framework, with fewer parameters and Flops, outperforms state-of-the-art methods on most metrics, demonstrating its superiority in occluded human pose estimation.

\begin{table}[htbp]
\footnotesize
	\caption{Comparison with SOTAs on the CrowdPose testing set.}\label{tab_sota_crowd}%
	\begin{tabular}{@{}lllllllll@{}}
		\toprule
		Methods & Backbone &  Input Size& Pa.(M)     & FL.(G)& $AP$ &  $AP^{E.}$ &  $AP^{M.}$ &  $AP^{H.}$  \\
  \midrule 
        KAPAO \cite{mcnally2022rethinking}                 &  KAPAO-M   &  1280x1280& 35.8 &  -   & 67.1 & 75.2 & 68.1 &  56.9 \\
        \tiny CSPNeXt\cite{lyu2022rtmdet} + Udp \cite{huang2020devil}                   &  CSPNeXt-m &  256x192  & 24.7 & 39.3 & 66.2 & 75.9 & 67.5 & 53.9  \\
		\midrule
        SimCC \cite{li2022simcc}                    &  HRNet-W32 &  256x192  & 31.3 & 7.1  & 66.7 & 74.1 & 67.8 & 56.2\\
        AdaptivePose \cite{xiao2022adaptivepose}             &  HRNet-W32 &  512x512  & 29.6 &  41.1   & 66.0 & 73.3 & 66.7 & {\bfseries 57.8} \\
        DEKR \cite{geng2021bottom}                     &  HRNet-W32 &  512x512  & 29.6 & 45.4 & 65.7 & 73.0 & 66.4 & 57.5 \\
        {\bfseries HRNet+LJOF}            & HRNet-W32  & 256x192 &28.5&7.1& {\bfseries 67.4} &	{\bfseries 76.7}	 & {\bfseries 68.6} &55.1\\	
        \midrule
        AdaptivePose \cite{xiao2022adaptivepose}             &  HRNet-W48 &  640x640  & 64.8 &  131.5   & 68.1 & 74.4 & 68.8 & {\bfseries 60.2} \\
        DEKR \cite{geng2021bottom}                     &  HRNet-W48 &  640x640	& 65.7 & 141.5& 67.3 & 74.6 & 68.1 & 58.7 	\\
        {\bfseries HRNet+LJOF}            & HRNet-W48  & 256x192 &63.6&14.6& {\bfseries 68.5} & {\bfseries 77.6}  & {\bfseries 69.7}	&56.9\\	
		\botrule
	\end{tabular}
\end{table}

\subsection{Ablation Study and Analysis}\label{subsec4.5}
This section aims to investigate the contribution of each proposed component in our framework to performance improvement, including limb joint augmentation and the dynamic structure loss in Section \ref{sec3}. We also analyze the influence of the parameters introduced by the proposed framework. In this section, all experiments use ViTPose-B as the baseline.

\textbf{Component analysis.} This paper first analyzes the effectiveness of each proposed component. The evaluation results on occluded datasets are shown in Table~\ref{tab4}. We can see that simply applying limb joint augmentation obtains a $1.0\%$ $AP$ gain and a $0.9\%$ $AR$ gain on the OCHuman testing set and a $0.4\%$ $AP$ gain on the CrowdPose testing set, indicating that the limb joint augmentation, which simulates the occluded scene of human joints in the image, is able to increase the robustness of the network to occluded samples. Adding DSL in Eq.\eqref{eq4} further boosts the performance gains to a $1.5\%$ $AP$ and a $1.3\%$ $AR$ on the OCHuman testing set and a $0.6\%$ $AP$ on the CrowdPose testing set. This demonstrates that the dynamic structure loss function is able to improve the network's ability to perceive the dependence between keypoints and modify the position of keypoints independently. Thus, it is concluded that generating occluded samples and learning structure information are important for occluded pose estimation.

\begin{table}[h]
	\caption{Analysis for limb joint augmentation and dynamic structure loss.}\label{tab4}%
	\begin{tabular}{@{}l|ll|llll@{}}
		\toprule
		\multirow{2}{*}{    }&\multicolumn{2}{c}{OCHuman test}&\multicolumn{4}{|c}{CrowdPose test}  \\ 
		& $AP$ & $AR$	&  $AP$ &$AP^{Easy}$  &$ AP^{Medium}$&$AP^{Hard}$    \\ 
		\botrule
		ViTPose-B   &59.5  & 63.9 & 66.5 &76.1 &67.9 &54.6   \\
		+LJA        & 60.5(\textbf{$\uparrow$ 1.0}) & 64.8(\textbf{$\uparrow$ 0.9}) & 66.9(\textbf{$\uparrow$ 0.4}) &76.4(\textbf{$\uparrow$ 0.3})	&68.2(\textbf{$\uparrow$ 0.3}) &55.1(\textbf{$\uparrow$ 0.5})   \\
		+DSL       & 61.0(\textbf{$\uparrow$ 1.5}) & 65.2(\textbf{$\uparrow$ 1.3}) & 67.1(\textbf{$\uparrow$ 0.6}) & 76.7(\textbf{$\uparrow$ 0.6})& 68.3(\textbf{$\uparrow$ 0.4}) &55.3(\textbf{$\uparrow$ 0.7})  \\		
		\botrule
	\end{tabular}
\end{table}

\textbf{The effect of the number and size of occlusion regions.} The number and size of occlusion regions are important hyper-parameters. Small occlusion regions are hard to provide enough occlusion information to the network, while large and more occlusion regions are more likely to damage the visual clues completely. Table \ref{tab5} tests different occlusion quantity ratios ${\alpha}$ with a range of $[0.15, 0.30]$ and various occlusion size ratios ${\beta}$ from $0.10$ to $0.20$ and reports their performance on OCHuman testing set sizes. From Table \ref{tab5}, it is evident that the highest $AP$ and $AR$ are achieved when ${\alpha}=0.15$, ${\beta=0.20}$. When the values of ${\alpha}$ and ${\beta}$ decrease, the generated training samples become easier, making the model less robust to occlusion scenarios. Conversely, increasing the values of ${\alpha}$ and ${\beta}$ results in an abundance of occluded areas in the training images, causing substantial damage to the visual information in the images.

\begin{table}[h]
	\caption{The effect of the number ratio ${\alpha}$ and size ratio ${\beta}$ of occlusion regions.}\label{tab5}%
	\begin{tabular}{@{}llllllll@{}}
		\toprule
		\multirow{2}{*}{Occlusion area ratio}&\multicolumn{2}{c}{OCHuman test}&\multicolumn{2}{c}{COCO val}    \\ 		
		& $AP$ & $AR$	& $AP$ & $AR$	     \\
		\botrule		
		\multicolumn{1}{l}{${\alpha}$=0.15,${\beta}$=0.10}  & 60.4 & 64.8 & 75.7 & 80.9   \\
		\multicolumn{1}{l}{${\alpha}$=0.15,${\beta}$=0.20}  & \textbf{60.5} & \textbf{64.8} & \textbf{75.8} & \textbf{81.1}   \\
		\multicolumn{1}{l}{${\alpha}$=0.20,${\beta}$=0.10}  & 60.3 & 64.7 & 75.7 & 80.9   \\
		\multicolumn{1}{l}{${\alpha}$=0.20,${\beta}$=0.20}  & 60.2 & 64.6 & 75.7 & 80.9   \\
		\multicolumn{1}{l}{${\alpha}$=0.25,${\beta}$=0.10}  & 60.4 & 64.7 & 75.7 & 81.0   \\
		\multicolumn{1}{l}{${\alpha}$=0.25,${\beta}$=0.20}  & 60.1 & 64.3 & 75.7 & 81.0   \\
		\multicolumn{1}{l}{${\alpha}$=0.30,${\beta}$=0.10}  & 60.1 & 64.5 & 75.6 & 80.9   \\
		\multicolumn{1}{l}{${\alpha}$=0.30,${\beta}$=0.20}  & 60.0 & 64.3 & 75.5 & 80.8  \\
		\botrule
	\end{tabular}
\end{table}

\textbf{Analysis of the weighting schemes in DSL.} To make the network focus on the localization of visible joints during the initial phase of training, this paper proposes to embed the structure constraint into the model training gradually. In this section, four weighting schemes are implemented to explore their influence on performance. The difference among the four weighting schemes has been illustrated in Fig. \ref{fig4}. Table \ref{tab7} shows the evaluation results on OCHuman of the weighting schemes. From Table \ref{tab7}, it is obvious that the step scheme achieves the largest performance gain, while the constant scheme obtains a slight performance improvement, demonstrating the effectiveness of the dynamic loss. Furthermore, it is worth noting that all the weighting schemes achieve better performance than the baseline with $\lambda=0$. It indicates that the structure loss is beneficial to the occluded pose estimation.

\begin{table}[h]\renewcommand\arraystretch{1.4}
	\caption{Analysis of the four weighting schemes in DSL}\label{tab7}%
	\begin{tabular}{@{}lllll@{}}
		\toprule
		\multirow{2}{*}{Structure  dynamic factor}&\multicolumn{2}{c}{OCHuman test} \\ 
		& $AP$ & $AR$	   \\ 
		\botrule	
		$\lambda_{0.00010}^{constant}$            & 60.6 & 64.8  \\
		$\lambda_{(0,0.00010)}^{linear}$          & 60.8 & 65.1   \\
		$\lambda_{(140,0,0.00010)}^{step}$        & \textbf{61.0} & \textbf{65.2}  \\
        $\lambda_{(0,0.00010)}^{exponential}$     & 60.6 & 64.9  \\
		\botrule
	\end{tabular}  
\end{table}

\textbf{Comparison with the existing structure losses.} Table \ref{tab_loss} shows the pose estimation results of existing structure losses on the OCHuman test set. All the training images are augmented by the proposed LJA for a fair comparison. In the experiments, we compare the proposed dynamic structural loss $L_{MSE}+\lambda L_{LSL}$ with the standard MSE loss $L_{MSE}$, the structure-aware loss $L_{MSE}+L_{SAL}$ \cite{ke2018multi}, and the structure-aware loss with dynamic weights $L_{MSE}+\lambda L_{SAL}$, where ``step'' scheme is utilized to weight the structure losses. From Table \ref{tab_loss}, it is observed that both structure losses with dynamic weights achieve performance gains compared to the standard MSE loss. It indicates that the prior knowledge of human body structure is beneficial to the occluded pose estimation. It is worth noting that adding the structure-aware loss from the beginning of training leads to performance deterioration, verifying the analysis about the weighting factor $\lambda$ in Section 3.3. Compared with the structure-aware loss with dynamic weights, the proposed dynamic structural loss boosts the performance gains to a $0.4\%$ AP and $0.2\%$ AR, demonstrating its superiority on occluded pose estimation.

\begin{table}[h]
	\centering\caption{Comparisons with existing structure losses on OCHuman test set.}\label{tab_loss}%
 \setlength{\tabcolsep}{7mm}{
	\begin{tabular}{lcc}
		\toprule
			Loss type&  $AP$ &$AR$      \\ 
		\midrule
  	$L_{MSE}$   &60.5 & 64.8       \\
   $L_{MSE}+L_{SAL}$  [22]  &    59.3 & 64.2 \\
	$L_{MSE}+\lambda L_{SAL}$ &    60.6 & 65.0 \\
        $L_{MSE}+\lambda L_{LSL}$ (Ours)  &   {\bfseries 61.0 }& {\bfseries 65.2}  \\	
        
		\bottomrule
	\end{tabular}}
\end{table}

\subsection{Visual Result Analysis}\label{subsec4.6}

\begin{figure}[htbp]
    \centering{\includegraphics[width=1.0\textwidth]{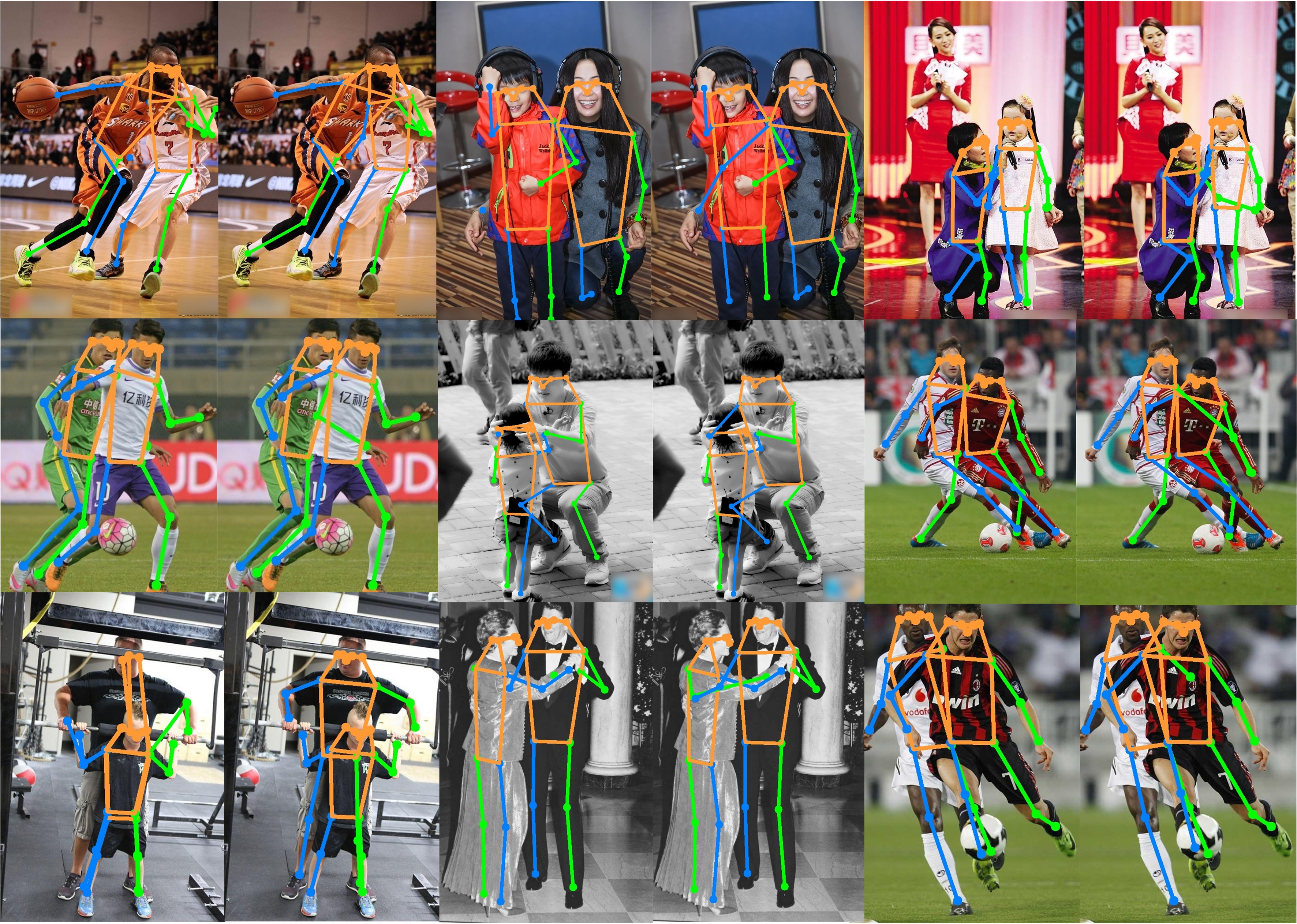}}
	\caption{The pose estimation results of the baseline ViTPose-B and the proposed framework on the occluded images from the OCHuman dataset. Please note that some human bodies do not have keypoints visualized because there are no labeled human boxes for them in the OCHuman dataset.\label{fig6}}
\end{figure}

Fig. \ref{fig6} illustrates the pose estimation results of the baseline ViTPose-B [8] and the proposed framework on the occluded image from the OCHuman dataset. It is worth noting that the human boxes in the OCHuman dataset are provided by annotations. Some human bodies in Fig. \ref{fig6}, such as the woman in the red clothing in the third set of the first row, are unannotated. As a result, the pose estimation model cannot process these bodies, and no keypoints are visualized for them.
The images on the left represent the predicted results of ViTPose-B. The evaluation results of the proposed framework are shown in the right columns. It is observed that the proposed framework performs better for occluded keypoints. For the second set of images in the first row, where the boy blocks the mother's right arm, the baseline network ViTPose-B cannot detect the lady's right elbow due to occlusion. However, the proposed method is still able to accurately predict the position of the blocked elbow. Similarly, in the second set of images of the third row, the man's right wrist is obscured. The baseline model fails to detect the occluded wrist, while the proposed method not only accurately locates the position of the man's right wrist, but also predicts the pose of the woman's left arm based on the dancing activity in the image.

\section{Conclusion}\label{sec5}

In this paper, we have proposed an occluded human pose estimation framework based on limb joint augmentation. The limb joint augmentation strategy is designed to simulate the real occluded scenes, generating limb joint-occlusion samples to train the pose estimation model. To leverage the prior knowledge of human body structure, a dynamic structure loss function is constructed based on limb graphs to explore the dependence between adjacent joints on human limbs. Compared with baselines, the proposed framework has achieved promising performance gains on two occlusion datasets.

\section*{Declarations}

\bmhead{Conflicts of interest} The authors have no conflicts of interest to declare that are relevant to the content of this article.

\bmhead{Data availability} The images in this study were taken from the public datasets. OCHuman, https://github.com/liruilong940607/OCHumanApi. CrowdPose, https://github.com/Jeff-sjtu/CrowdPose. Coco, https://cocodataset.org/\#home.


\bibliography{occlusionHPE-bibliography}


\begin{thebibliography}{47}
\ifx \bisbn   \undefined \def \bisbn  #1{ISBN #1}\fi
\ifx \binits  \undefined \def \binits#1{#1}\fi
\ifx \bauthor  \undefined \def \bauthor#1{#1}\fi
\ifx \batitle  \undefined \def \batitle#1{#1}\fi
\ifx \bjtitle  \undefined \def \bjtitle#1{#1}\fi
\ifx \bvolume  \undefined \def \bvolume#1{\textbf{#1}}\fi
\ifx \byear  \undefined \def \byear#1{#1}\fi
\ifx \bissue  \undefined \def \bissue#1{#1}\fi
\ifx \bfpage  \undefined \def \bfpage#1{#1}\fi
\ifx \blpage  \undefined \def \blpage #1{#1}\fi
\ifx \burl  \undefined \def \burl#1{\textsf{#1}}\fi
\ifx \doiurl  \undefined \def \doiurl#1{\url{https://doi.org/#1}}\fi
\ifx \betal  \undefined \def \betal{\textit{et al.}}\fi
\ifx \binstitute  \undefined \def \binstitute#1{#1}\fi
\ifx \binstitutionaled  \undefined \def \binstitutionaled#1{#1}\fi
\ifx \bctitle  \undefined \def \bctitle#1{#1}\fi
\ifx \beditor  \undefined \def \beditor#1{#1}\fi
\ifx \bpublisher  \undefined \def \bpublisher#1{#1}\fi
\ifx \bbtitle  \undefined \def \bbtitle#1{#1}\fi
\ifx \bedition  \undefined \def \bedition#1{#1}\fi
\ifx \bseriesno  \undefined \def \bseriesno#1{#1}\fi
\ifx \blocation  \undefined \def \blocation#1{#1}\fi
\ifx \bsertitle  \undefined \def \bsertitle#1{#1}\fi
\ifx \bsnm \undefined \def \bsnm#1{#1}\fi
\ifx \bsuffix \undefined \def \bsuffix#1{#1}\fi
\ifx \bparticle \undefined \def \bparticle#1{#1}\fi
\ifx \barticle \undefined \def \barticle#1{#1}\fi
\bibcommenthead
\ifx \bconfdate \undefined \def \bconfdate #1{#1}\fi
\ifx \botherref \undefined \def \botherref #1{#1}\fi
\ifx \url \undefined \def \url#1{\textsf{#1}}\fi
\ifx \bchapter \undefined \def \bchapter#1{#1}\fi
\ifx \bbook \undefined \def \bbook#1{#1}\fi
\ifx \bcomment \undefined \def \bcomment#1{#1}\fi
\ifx \oauthor \undefined \def \oauthor#1{#1}\fi
\ifx \citeauthoryear \undefined \def \citeauthoryear#1{#1}\fi
\ifx \endbibitem  \undefined \def \endbibitem {}\fi
\ifx \bconflocation  \undefined \def \bconflocation#1{#1}\fi
\ifx \arxivurl  \undefined \def \arxivurl#1{\textsf{#1}}\fi
\csname PreBibitemsHook\endcsname

\bibitem[\protect\citeauthoryear{Fu et~al.}{2023}]{bib3}
\begin{botherref}
\oauthor{\bsnm{Fu}, \binits{Y.}},
\oauthor{\bsnm{Meng}, \binits{S.}},
\oauthor{\bsnm{Hou}, \binits{S.}},
\oauthor{\bsnm{Hu}, \binits{X.}},
\oauthor{\bsnm{Huang}, \binits{Y.}}:
Gpgait: Generalized pose-based gait recognition.
arXiv preprint arXiv:2303.05234
(2023)
\end{botherref}
\endbibitem

\bibitem[\protect\citeauthoryear{Park et~al.}{2023}]{bib2}
\begin{bchapter}
\bauthor{\bsnm{Park}, \binits{J.}},
\bauthor{\bsnm{Park}, \binits{J.-W.}},
\bauthor{\bsnm{Lee}, \binits{J.-S.}}:
\bctitle{Viplo: Vision transformer based pose-conditioned self-loop graph for human-object interaction detection}.
In: \bbtitle{Proceedings of the IEEE/CVF Conference on Computer Vision and Pattern Recognition},
pp. \bfpage{17152}--\blpage{17162}
(\byear{2023})
\end{bchapter}
\endbibitem

\bibitem[\protect\citeauthoryear{Azadi et~al.}{2023}]{bib1}
\begin{botherref}
\oauthor{\bsnm{Azadi}, \binits{S.}},
\oauthor{\bsnm{Shah}, \binits{A.}},
\oauthor{\bsnm{Hayes}, \binits{T.}},
\oauthor{\bsnm{Parikh}, \binits{D.}},
\oauthor{\bsnm{Gupta}, \binits{S.}}:
Make-an-animation: Large-scale text-conditional 3d human motion generation.
arXiv preprint arXiv:2305.09662
(2023)
\end{botherref}
\endbibitem

\bibitem[\protect\citeauthoryear{Newell et~al.}{2016}]{newell2016stacked}
\begin{bchapter}
\bauthor{\bsnm{Newell}, \binits{A.}},
\bauthor{\bsnm{Yang}, \binits{K.}},
\bauthor{\bsnm{Deng}, \binits{J.}}:
\bctitle{Stacked hourglass networks for human pose estimation}.
In: \bbtitle{Computer Vision--ECCV 2016: 14th European Conference, Amsterdam, The Netherlands, October 11-14, 2016, Proceedings, Part VIII 14},
pp. \bfpage{483}--\blpage{499}
(\byear{2016}).
\bcomment{Springer}
\end{bchapter}
\endbibitem

\bibitem[\protect\citeauthoryear{Chu et~al.}{2017}]{chu2017multi}
\begin{bchapter}
\bauthor{\bsnm{Chu}, \binits{X.}},
\bauthor{\bsnm{Yang}, \binits{W.}},
\bauthor{\bsnm{Ouyang}, \binits{W.}},
\bauthor{\bsnm{Ma}, \binits{C.}},
\bauthor{\bsnm{Yuille}, \binits{A.L.}},
\bauthor{\bsnm{Wang}, \binits{X.}}:
\bctitle{Multi-context attention for human pose estimation}.
In: \bbtitle{Proceedings of the IEEE Conference on Computer Vision and Pattern Recognition},
pp. \bfpage{1831}--\blpage{1840}
(\byear{2017})
\end{bchapter}
\endbibitem

\bibitem[\protect\citeauthoryear{Yang et~al.}{2021}]{bib14}
\begin{bchapter}
\bauthor{\bsnm{Yang}, \binits{S.}},
\bauthor{\bsnm{Quan}, \binits{Z.}},
\bauthor{\bsnm{Nie}, \binits{M.}},
\bauthor{\bsnm{Yang}, \binits{W.}}:
\bctitle{Transpose: Keypoint localization via transformer}.
In: \bbtitle{Proceedings of the IEEE/CVF International Conference on Computer Vision},
pp. \bfpage{11802}--\blpage{11812}
(\byear{2021})
\end{bchapter}
\endbibitem

\bibitem[\protect\citeauthoryear{Jiang et~al.}{2023}]{bib15}
\begin{botherref}
\oauthor{\bsnm{Jiang}, \binits{T.}},
\oauthor{\bsnm{Lu}, \binits{P.}},
\oauthor{\bsnm{Zhang}, \binits{L.}},
\oauthor{\bsnm{Ma}, \binits{N.}},
\oauthor{\bsnm{Han}, \binits{R.}},
\oauthor{\bsnm{Lyu}, \binits{C.}},
\oauthor{\bsnm{Li}, \binits{Y.}},
\oauthor{\bsnm{Chen}, \binits{K.}}:
Rtmpose: Real-time multi-person pose estimation based on mmpose.
arXiv preprint arXiv:2303.07399
(2023)
\end{botherref}
\endbibitem

\bibitem[\protect\citeauthoryear{Xu et~al.}{2024}]{xu2024disentangled}
\begin{botherref}
\oauthor{\bsnm{Xu}, \binits{W.}},
\oauthor{\bsnm{Long}, \binits{C.}},
\oauthor{\bsnm{Nie}, \binits{Y.}},
\oauthor{\bsnm{Wang}, \binits{G.}}:
Disentangled representation learning for controllable person image generation.
IEEE Transactions on Multimedia
(2024)
\end{botherref}
\endbibitem

\bibitem[\protect\citeauthoryear{Khirodkar et~al.}{2021}]{khirodkar2021multi}
\begin{bchapter}
\bauthor{\bsnm{Khirodkar}, \binits{R.}},
\bauthor{\bsnm{Chari}, \binits{V.}},
\bauthor{\bsnm{Agrawal}, \binits{A.}},
\bauthor{\bsnm{Tyagi}, \binits{A.}}:
\bctitle{Multi-instance pose networks: Rethinking top-down pose estimation}.
In: \bbtitle{Proceedings of the IEEE/CVF International Conference on Computer Vision},
pp. \bfpage{3122}--\blpage{3131}
(\byear{2021})
\end{bchapter}
\endbibitem

\bibitem[\protect\citeauthoryear{Peng et~al.}{2018}]{peng2018jointly}
\begin{bchapter}
\bauthor{\bsnm{Peng}, \binits{X.}},
\bauthor{\bsnm{Tang}, \binits{Z.}},
\bauthor{\bsnm{Yang}, \binits{F.}},
\bauthor{\bsnm{Feris}, \binits{R.S.}},
\bauthor{\bsnm{Metaxas}, \binits{D.}}:
\bctitle{Jointly optimize data augmentation and network training: Adversarial data augmentation in human pose estimation}.
In: \bbtitle{Proceedings of the IEEE Conference on Computer Vision and Pattern Recognition},
pp. \bfpage{2226}--\blpage{2234}
(\byear{2018})
\end{bchapter}
\endbibitem

\bibitem[\protect\citeauthoryear{Iqbal and Gall}{2016}]{iqbal2016multi}
\begin{bchapter}
\bauthor{\bsnm{Iqbal}, \binits{U.}},
\bauthor{\bsnm{Gall}, \binits{J.}}:
\bctitle{Multi-person pose estimation with local joint-to-person associations}.
In: \bbtitle{Computer Vision--ECCV 2016 Workshops: Amsterdam, The Netherlands, October 8-10 and 15-16, 2016, Proceedings, Part II 14},
pp. \bfpage{627}--\blpage{642}
(\byear{2016}).
\bcomment{Springer}
\end{bchapter}
\endbibitem

\bibitem[\protect\citeauthoryear{Chen et~al.}{2018}]{chen2018cascaded}
\begin{bchapter}
\bauthor{\bsnm{Chen}, \binits{Y.}},
\bauthor{\bsnm{Wang}, \binits{Z.}},
\bauthor{\bsnm{Peng}, \binits{Y.}},
\bauthor{\bsnm{Zhang}, \binits{Z.}},
\bauthor{\bsnm{Yu}, \binits{G.}},
\bauthor{\bsnm{Sun}, \binits{J.}}:
\bctitle{Cascaded pyramid network for multi-person pose estimation}.
In: \bbtitle{Proceedings of the IEEE Conference on Computer Vision and Pattern Recognition},
pp. \bfpage{7103}--\blpage{7112}
(\byear{2018})
\end{bchapter}
\endbibitem

\bibitem[\protect\citeauthoryear{Su et~al.}{2019}]{su2019multi}
\begin{bchapter}
\bauthor{\bsnm{Su}, \binits{K.}},
\bauthor{\bsnm{Yu}, \binits{D.}},
\bauthor{\bsnm{Xu}, \binits{Z.}},
\bauthor{\bsnm{Geng}, \binits{X.}},
\bauthor{\bsnm{Wang}, \binits{C.}}:
\bctitle{Multi-person pose estimation with enhanced channel-wise and spatial information}.
In: \bbtitle{Proceedings of the IEEE/CVF Conference on Computer Vision and Pattern Recognition},
pp. \bfpage{5674}--\blpage{5682}
(\byear{2019})
\end{bchapter}
\endbibitem

\bibitem[\protect\citeauthoryear{Li et~al.}{2021}]{li2021tokenpose}
\begin{bchapter}
\bauthor{\bsnm{Li}, \binits{Y.}},
\bauthor{\bsnm{Zhang}, \binits{S.}},
\bauthor{\bsnm{Wang}, \binits{Z.}},
\bauthor{\bsnm{Yang}, \binits{S.}},
\bauthor{\bsnm{Yang}, \binits{W.}},
\bauthor{\bsnm{Xia}, \binits{S.-T.}},
\bauthor{\bsnm{Zhou}, \binits{E.}}:
\bctitle{Tokenpose: Learning keypoint tokens for human pose estimation}.
In: \bbtitle{Proceedings of the IEEE/CVF International Conference on Computer Vision},
pp. \bfpage{11313}--\blpage{11322}
(\byear{2021})
\end{bchapter}
\endbibitem

\bibitem[\protect\citeauthoryear{Ma et~al.}{2022}]{ma2022ppt}
\begin{bchapter}
\bauthor{\bsnm{Ma}, \binits{H.}},
\bauthor{\bsnm{Wang}, \binits{Z.}},
\bauthor{\bsnm{Chen}, \binits{Y.}},
\bauthor{\bsnm{Kong}, \binits{D.}},
\bauthor{\bsnm{Chen}, \binits{L.}},
\bauthor{\bsnm{Liu}, \binits{X.}},
\bauthor{\bsnm{Yan}, \binits{X.}},
\bauthor{\bsnm{Tang}, \binits{H.}},
\bauthor{\bsnm{Xie}, \binits{X.}}:
\bctitle{Ppt: token-pruned pose transformer for monocular and multi-view human pose estimation}.
In: \bbtitle{European Conference on Computer Vision},
pp. \bfpage{424}--\blpage{442}
(\byear{2022}).
\bcomment{Springer}
\end{bchapter}
\endbibitem

\bibitem[\protect\citeauthoryear{Shi et~al.}{2022}]{shi2022end}
\begin{bchapter}
\bauthor{\bsnm{Shi}, \binits{D.}},
\bauthor{\bsnm{Wei}, \binits{X.}},
\bauthor{\bsnm{Li}, \binits{L.}},
\bauthor{\bsnm{Ren}, \binits{Y.}},
\bauthor{\bsnm{Tan}, \binits{W.}}:
\bctitle{End-to-end multi-person pose estimation with transformers}.
In: \bbtitle{Proceedings of the IEEE/CVF Conference on Computer Vision and Pattern Recognition},
pp. \bfpage{11069}--\blpage{11078}
(\byear{2022})
\end{bchapter}
\endbibitem

\bibitem[\protect\citeauthoryear{Zheng et~al.}{2023}]{zheng2023deep}
\begin{barticle}
\bauthor{\bsnm{Zheng}, \binits{C.}},
\bauthor{\bsnm{Wu}, \binits{W.}},
\bauthor{\bsnm{Chen}, \binits{C.}},
\bauthor{\bsnm{Yang}, \binits{T.}},
\bauthor{\bsnm{Zhu}, \binits{S.}},
\bauthor{\bsnm{Shen}, \binits{J.}},
\bauthor{\bsnm{Kehtarnavaz}, \binits{N.}},
\bauthor{\bsnm{Shah}, \binits{M.}}:
\batitle{Deep learning-based human pose estimation: A survey}.
\bjtitle{ACM Computing Surveys}
\bvolume{56}(\bissue{1}),
\bfpage{1}--\blpage{37}
(\byear{2023})
\end{barticle}
\endbibitem

\bibitem[\protect\citeauthoryear{Wei et~al.}{2016}]{wei2016convolutional}
\begin{bchapter}
\bauthor{\bsnm{Wei}, \binits{S.-E.}},
\bauthor{\bsnm{Ramakrishna}, \binits{V.}},
\bauthor{\bsnm{Kanade}, \binits{T.}},
\bauthor{\bsnm{Sheikh}, \binits{Y.}}:
\bctitle{Convolutional pose machines}.
In: \bbtitle{Proceedings of the IEEE Conference on Computer Vision and Pattern Recognition},
pp. \bfpage{4724}--\blpage{4732}
(\byear{2016})
\end{bchapter}
\endbibitem

\bibitem[\protect\citeauthoryear{Sun et~al.}{2019}]{bib8}
\begin{bchapter}
\bauthor{\bsnm{Sun}, \binits{K.}},
\bauthor{\bsnm{Xiao}, \binits{B.}},
\bauthor{\bsnm{Liu}, \binits{D.}},
\bauthor{\bsnm{Wang}, \binits{J.}}:
\bctitle{Deep high-resolution representation learning for human pose estimation}.
In: \bbtitle{Proceedings of the IEEE/CVF Conference on Computer Vision and Pattern Recognition},
pp. \bfpage{5693}--\blpage{5703}
(\byear{2019})
\end{bchapter}
\endbibitem

\bibitem[\protect\citeauthoryear{Yuan et~al.}{2021}]{bib13}
\begin{botherref}
\oauthor{\bsnm{Yuan}, \binits{Y.}},
\oauthor{\bsnm{Fu}, \binits{R.}},
\oauthor{\bsnm{Huang}, \binits{L.}},
\oauthor{\bsnm{Lin}, \binits{W.}},
\oauthor{\bsnm{Zhang}, \binits{C.}},
\oauthor{\bsnm{Chen}, \binits{X.}},
\oauthor{\bsnm{Wang}, \binits{J.}}:
Hrformer: High-resolution transformer for dense prediction.
arXiv preprint arXiv:2110.09408
(2021)
\end{botherref}
\endbibitem

\bibitem[\protect\citeauthoryear{Chu et~al.}{2016}]{chu2016structured}
\begin{bchapter}
\bauthor{\bsnm{Chu}, \binits{X.}},
\bauthor{\bsnm{Ouyang}, \binits{W.}},
\bauthor{\bsnm{Li}, \binits{H.}},
\bauthor{\bsnm{Wang}, \binits{X.}}:
\bctitle{Structured feature learning for pose estimation}.
In: \bbtitle{Proceedings of the IEEE Conference on Computer Vision and Pattern Recognition},
pp. \bfpage{4715}--\blpage{4723}
(\byear{2016})
\end{bchapter}
\endbibitem

\bibitem[\protect\citeauthoryear{Ke et~al.}{2018}]{ke2018multi}
\begin{bchapter}
\bauthor{\bsnm{Ke}, \binits{L.}},
\bauthor{\bsnm{Chang}, \binits{M.-C.}},
\bauthor{\bsnm{Qi}, \binits{H.}},
\bauthor{\bsnm{Lyu}, \binits{S.}}:
\bctitle{Multi-scale structure-aware network for human pose estimation}.
In: \bbtitle{Proceedings of the European Conference on Computer Vision (ECCV)},
pp. \bfpage{713}--\blpage{728}
(\byear{2018})
\end{bchapter}
\endbibitem

\bibitem[\protect\citeauthoryear{Tang et~al.}{2018}]{tang2018deeply}
\begin{bchapter}
\bauthor{\bsnm{Tang}, \binits{W.}},
\bauthor{\bsnm{Yu}, \binits{P.}},
\bauthor{\bsnm{Wu}, \binits{Y.}}:
\bctitle{Deeply learned compositional models for human pose estimation}.
In: \bbtitle{Proceedings of the European Conference on Computer Vision (ECCV)},
pp. \bfpage{190}--\blpage{206}
(\byear{2018})
\end{bchapter}
\endbibitem

\bibitem[\protect\citeauthoryear{Pishchulin et~al.}{2016}]{pishchulin2016deepcut}
\begin{bchapter}
\bauthor{\bsnm{Pishchulin}, \binits{L.}},
\bauthor{\bsnm{Insafutdinov}, \binits{E.}},
\bauthor{\bsnm{Tang}, \binits{S.}},
\bauthor{\bsnm{Andres}, \binits{B.}},
\bauthor{\bsnm{Andriluka}, \binits{M.}},
\bauthor{\bsnm{Gehler}, \binits{P.V.}},
\bauthor{\bsnm{Schiele}, \binits{B.}}:
\bctitle{Deepcut: Joint subset partition and labeling for multi person pose estimation}.
In: \bbtitle{Proceedings of the IEEE Conference on Computer Vision and Pattern Recognition},
pp. \bfpage{4929}--\blpage{4937}
(\byear{2016})
\end{bchapter}
\endbibitem

\bibitem[\protect\citeauthoryear{Cao et~al.}{2017}]{cao2017realtime}
\begin{bchapter}
\bauthor{\bsnm{Cao}, \binits{Z.}},
\bauthor{\bsnm{Simon}, \binits{T.}},
\bauthor{\bsnm{Wei}, \binits{S.-E.}},
\bauthor{\bsnm{Sheikh}, \binits{Y.}}:
\bctitle{Realtime multi-person 2d pose estimation using part affinity fields}.
In: \bbtitle{Proceedings of the IEEE Conference on Computer Vision and Pattern Recognition},
pp. \bfpage{7291}--\blpage{7299}
(\byear{2017})
\end{bchapter}
\endbibitem

\bibitem[\protect\citeauthoryear{Cheng et~al.}{2020}]{bib7}
\begin{bchapter}
\bauthor{\bsnm{Cheng}, \binits{B.}},
\bauthor{\bsnm{Xiao}, \binits{B.}},
\bauthor{\bsnm{Wang}, \binits{J.}},
\bauthor{\bsnm{Shi}, \binits{H.}},
\bauthor{\bsnm{Huang}, \binits{T.S.}},
\bauthor{\bsnm{Zhang}, \binits{L.}}:
\bctitle{Higherhrnet: Scale-aware representation learning for bottom-up human pose estimation}.
In: \bbtitle{Proceedings of the IEEE/CVF Conference on Computer Vision and Pattern Recognition},
pp. \bfpage{5386}--\blpage{5395}
(\byear{2020})
\end{bchapter}
\endbibitem

\bibitem[\protect\citeauthoryear{Wang et~al.}{2022}]{bib9}
\begin{bchapter}
\bauthor{\bsnm{Wang}, \binits{Y.}},
\bauthor{\bsnm{Li}, \binits{M.}},
\bauthor{\bsnm{Cai}, \binits{H.}},
\bauthor{\bsnm{Chen}, \binits{W.-M.}},
\bauthor{\bsnm{Han}, \binits{S.}}:
\bctitle{Lite pose: Efficient architecture design for 2d human pose estimation}.
In: \bbtitle{Proceedings of the IEEE/CVF Conference on Computer Vision and Pattern Recognition},
pp. \bfpage{13126}--\blpage{13136}
(\byear{2022})
\end{bchapter}
\endbibitem

\bibitem[\protect\citeauthoryear{Wang et~al.}{2023}]{bib10}
\begin{bchapter}
\bauthor{\bsnm{Wang}, \binits{H.}},
\bauthor{\bsnm{Liu}, \binits{J.}},
\bauthor{\bsnm{Tang}, \binits{J.}},
\bauthor{\bsnm{Wu}, \binits{G.}}:
\bctitle{Lightweight super-resolution head for human pose estimation}.
In: \bbtitle{Proceedings of the 31st ACM International Conference on Multimedia},
pp. \bfpage{2353}--\blpage{2361}
(\byear{2023})
\end{bchapter}
\endbibitem

\bibitem[\protect\citeauthoryear{Zhang et~al.}{2019}]{bib35}
\begin{bchapter}
\bauthor{\bsnm{Zhang}, \binits{S.-H.}},
\bauthor{\bsnm{Li}, \binits{R.}},
\bauthor{\bsnm{Dong}, \binits{X.}},
\bauthor{\bsnm{Rosin}, \binits{P.}},
\bauthor{\bsnm{Cai}, \binits{Z.}},
\bauthor{\bsnm{Han}, \binits{X.}},
\bauthor{\bsnm{Yang}, \binits{D.}},
\bauthor{\bsnm{Huang}, \binits{H.}},
\bauthor{\bsnm{Hu}, \binits{S.-M.}}:
\bctitle{Pose2seg: Detection free human instance segmentation}.
In: \bbtitle{Proceedings of the IEEE/CVF Conference on Computer Vision and Pattern Recognition},
pp. \bfpage{889}--\blpage{898}
(\byear{2019})
\end{bchapter}
\endbibitem

\bibitem[\protect\citeauthoryear{Li et~al.}{2019}]{bib36}
\begin{bchapter}
\bauthor{\bsnm{Li}, \binits{J.}},
\bauthor{\bsnm{Wang}, \binits{C.}},
\bauthor{\bsnm{Zhu}, \binits{H.}},
\bauthor{\bsnm{Mao}, \binits{Y.}},
\bauthor{\bsnm{Fang}, \binits{H.-S.}},
\bauthor{\bsnm{Lu}, \binits{C.}}:
\bctitle{Crowdpose: Efficient crowded scenes pose estimation and a new benchmark}.
In: \bbtitle{Proceedings of the IEEE/CVF Conference on Computer Vision and Pattern Recognition},
pp. \bfpage{10863}--\blpage{10872}
(\byear{2019})
\end{bchapter}
\endbibitem

\bibitem[\protect\citeauthoryear{Zhang et~al.}{2021}]{zhang2021six}
\begin{bchapter}
\bauthor{\bsnm{Zhang}, \binits{T.}},
\bauthor{\bsnm{Ma}, \binits{W.}},
\bauthor{\bsnm{Wang}, \binits{G.}}:
\bctitle{Six-channel image representation for cross-domain object detection}.
In: \bbtitle{Image and Graphics: 11th International Conference, ICIG 2021, Haikou, China, August 6--8, 2021, Proceedings, Part I 11},
pp. \bfpage{171}--\blpage{184}
(\byear{2021}).
\bcomment{Springer}
\end{bchapter}
\endbibitem

\bibitem[\protect\citeauthoryear{Li et~al.}{2020}]{li2020cascaded}
\begin{bchapter}
\bauthor{\bsnm{Li}, \binits{S.}},
\bauthor{\bsnm{Ke}, \binits{L.}},
\bauthor{\bsnm{Pratama}, \binits{K.}},
\bauthor{\bsnm{Tai}, \binits{Y.-W.}},
\bauthor{\bsnm{Tang}, \binits{C.-K.}},
\bauthor{\bsnm{Cheng}, \binits{K.-T.}}:
\bctitle{Cascaded deep monocular 3d human pose estimation with evolutionary training data}.
In: \bbtitle{Proceedings of the IEEE/CVF Conference on Computer Vision and Pattern Recognition},
pp. \bfpage{6173}--\blpage{6183}
(\byear{2020})
\end{bchapter}
\endbibitem

\bibitem[\protect\citeauthoryear{Xu and Wang}{2021}]{xu2021domain}
\begin{barticle}
\bauthor{\bsnm{Xu}, \binits{W.}},
\bauthor{\bsnm{Wang}, \binits{G.}}:
\batitle{A domain gap aware generative adversarial network for multi-domain image translation}.
\bjtitle{IEEE Transactions on Image Processing}
\bvolume{31},
\bfpage{72}--\blpage{84}
(\byear{2021})
\end{barticle}
\endbibitem

\bibitem[\protect\citeauthoryear{Tompson et~al.}{2014}]{bib33}
\begin{botherref}
\oauthor{\bsnm{Tompson}, \binits{J.J.}},
\oauthor{\bsnm{Jain}, \binits{A.}},
\oauthor{\bsnm{LeCun}, \binits{Y.}},
\oauthor{\bsnm{Bregler}, \binits{C.}}:
Joint training of a convolutional network and a graphical model for human pose estimation.
Advances in neural information processing systems
\textbf{27}
(2014)
\end{botherref}
\endbibitem

\bibitem[\protect\citeauthoryear{Lin et~al.}{2014}]{bib34}
\begin{bchapter}
\bauthor{\bsnm{Lin}, \binits{T.-Y.}},
\bauthor{\bsnm{Maire}, \binits{M.}},
\bauthor{\bsnm{Belongie}, \binits{S.}},
\bauthor{\bsnm{Hays}, \binits{J.}},
\bauthor{\bsnm{Perona}, \binits{P.}},
\bauthor{\bsnm{Ramanan}, \binits{D.}},
\bauthor{\bsnm{Doll{\'a}r}, \binits{P.}},
\bauthor{\bsnm{Zitnick}, \binits{C.L.}}:
\bctitle{Microsoft coco: Common objects in context}.
In: \bbtitle{Computer Vision--ECCV 2014: 13th European Conference, Zurich, Switzerland, September 6-12, 2014, Proceedings, Part V 13},
pp. \bfpage{740}--\blpage{755}
(\byear{2014}).
\bcomment{Springer}
\end{bchapter}
\endbibitem

\bibitem[\protect\citeauthoryear{Geng et~al.}{2021}]{bib17}
\begin{bchapter}
\bauthor{\bsnm{Geng}, \binits{Z.}},
\bauthor{\bsnm{Sun}, \binits{K.}},
\bauthor{\bsnm{Xiao}, \binits{B.}},
\bauthor{\bsnm{Zhang}, \binits{Z.}},
\bauthor{\bsnm{Wang}, \binits{J.}}:
\bctitle{Bottom-up human pose estimation via disentangled keypoint regression}.
In: \bbtitle{Proceedings of the IEEE/CVF Conference on Computer Vision and Pattern Recognition},
pp. \bfpage{14676}--\blpage{14686}
(\byear{2021})
\end{bchapter}
\endbibitem

\bibitem[\protect\citeauthoryear{Xiao et~al.}{2018}]{bib11}
\begin{bchapter}
\bauthor{\bsnm{Xiao}, \binits{B.}},
\bauthor{\bsnm{Wu}, \binits{H.}},
\bauthor{\bsnm{Wei}, \binits{Y.}}:
\bctitle{Simple baselines for human pose estimation and tracking}.
In: \bbtitle{Proceedings of the European Conference on Computer Vision (ECCV)},
pp. \bfpage{466}--\blpage{481}
(\byear{2018})
\end{bchapter}
\endbibitem

\bibitem[\protect\citeauthoryear{Xu et~al.}{2022}]{bib16}
\begin{barticle}
\bauthor{\bsnm{Xu}, \binits{Y.}},
\bauthor{\bsnm{Zhang}, \binits{J.}},
\bauthor{\bsnm{Zhang}, \binits{Q.}},
\bauthor{\bsnm{Tao}, \binits{D.}}:
\batitle{Vitpose: Simple vision transformer baselines for human pose estimation}.
\bjtitle{Advances in Neural Information Processing Systems}
\bvolume{35},
\bfpage{38571}--\blpage{38584}
(\byear{2022})
\end{barticle}
\endbibitem

\bibitem[\protect\citeauthoryear{Wang et~al.}{2023}]{wang2023lightweight}
\begin{bchapter}
\bauthor{\bsnm{Wang}, \binits{H.}},
\bauthor{\bsnm{Liu}, \binits{J.}},
\bauthor{\bsnm{Tang}, \binits{J.}},
\bauthor{\bsnm{Wu}, \binits{G.}}:
\bctitle{Lightweight super-resolution head for human pose estimation}.
In: \bbtitle{Proceedings of the 31st ACM International Conference on Multimedia},
pp. \bfpage{2353}--\blpage{2361}
(\byear{2023})
\end{bchapter}
\endbibitem

\bibitem[\protect\citeauthoryear{Wang et~al.}{2021}]{wang2021robust}
\begin{barticle}
\bauthor{\bsnm{Wang}, \binits{D.}},
\bauthor{\bsnm{Zhang}, \binits{S.}},
\bauthor{\bsnm{Hua}, \binits{G.}}:
\batitle{Robust pose estimation in crowded scenes with direct pose-level inference}.
\bjtitle{Advances in Neural Information Processing Systems}
\bvolume{34},
\bfpage{6278}--\blpage{6289}
(\byear{2021})
\end{barticle}
\endbibitem

\bibitem[\protect\citeauthoryear{Jeong et~al.}{2023}]{jeong2023boir}
\begin{botherref}
\oauthor{\bsnm{Jeong}, \binits{U.}},
\oauthor{\bsnm{Baek}, \binits{S.}},
\oauthor{\bsnm{Chang}, \binits{H.J.}},
\oauthor{\bsnm{Kim}, \binits{K.I.}}:
Boir: Box-supervised instance representation for multi-person pose estimation.
arXiv preprint arXiv:2309.14072
(2023)
\end{botherref}
\endbibitem

\bibitem[\protect\citeauthoryear{McNally et~al.}{2022}]{mcnally2022rethinking}
\begin{bchapter}
\bauthor{\bsnm{McNally}, \binits{W.}},
\bauthor{\bsnm{Vats}, \binits{K.}},
\bauthor{\bsnm{Wong}, \binits{A.}},
\bauthor{\bsnm{McPhee}, \binits{J.}}:
\bctitle{Rethinking keypoint representations: Modeling keypoints and poses as objects for multi-person human pose estimation}.
In: \bbtitle{European Conference on Computer Vision},
pp. \bfpage{37}--\blpage{54}
(\byear{2022}).
\bcomment{Springer}
\end{bchapter}
\endbibitem

\bibitem[\protect\citeauthoryear{Lyu et~al.}{2022}]{lyu2022rtmdet}
\begin{botherref}
\oauthor{\bsnm{Lyu}, \binits{C.}},
\oauthor{\bsnm{Zhang}, \binits{W.}},
\oauthor{\bsnm{Huang}, \binits{H.}},
\oauthor{\bsnm{Zhou}, \binits{Y.}},
\oauthor{\bsnm{Wang}, \binits{Y.}},
\oauthor{\bsnm{Liu}, \binits{Y.}},
\oauthor{\bsnm{Zhang}, \binits{S.}},
\oauthor{\bsnm{Chen}, \binits{K.}}:
Rtmdet: An empirical study of designing real-time object detectors.
arXiv preprint arXiv:2212.07784
(2022)
\end{botherref}
\endbibitem

\bibitem[\protect\citeauthoryear{Huang et~al.}{2020}]{huang2020devil}
\begin{bchapter}
\bauthor{\bsnm{Huang}, \binits{J.}},
\bauthor{\bsnm{Zhu}, \binits{Z.}},
\bauthor{\bsnm{Guo}, \binits{F.}},
\bauthor{\bsnm{Huang}, \binits{G.}}:
\bctitle{The devil is in the details: Delving into unbiased data processing for human pose estimation}.
In: \bbtitle{Proceedings of the IEEE/CVF Conference on Computer Vision and Pattern Recognition},
pp. \bfpage{5700}--\blpage{5709}
(\byear{2020})
\end{bchapter}
\endbibitem

\bibitem[\protect\citeauthoryear{Li et~al.}{2022}]{li2022simcc}
\begin{bchapter}
\bauthor{\bsnm{Li}, \binits{Y.}},
\bauthor{\bsnm{Yang}, \binits{S.}},
\bauthor{\bsnm{Liu}, \binits{P.}},
\bauthor{\bsnm{Zhang}, \binits{S.}},
\bauthor{\bsnm{Wang}, \binits{Y.}},
\bauthor{\bsnm{Wang}, \binits{Z.}},
\bauthor{\bsnm{Yang}, \binits{W.}},
\bauthor{\bsnm{Xia}, \binits{S.-T.}}:
\bctitle{Simcc: A simple coordinate classification perspective for human pose estimation}.
In: \bbtitle{European Conference on Computer Vision},
pp. \bfpage{89}--\blpage{106}
(\byear{2022}).
\bcomment{Springer}
\end{bchapter}
\endbibitem

\bibitem[\protect\citeauthoryear{Xiao et~al.}{2022}]{xiao2022adaptivepose}
\begin{bchapter}
\bauthor{\bsnm{Xiao}, \binits{Y.}},
\bauthor{\bsnm{Wang}, \binits{X.J.}},
\bauthor{\bsnm{Yu}, \binits{D.}},
\bauthor{\bsnm{Wang}, \binits{G.}},
\bauthor{\bsnm{Zhang}, \binits{Q.}},
\bauthor{\bsnm{Mingshu}, \binits{H.}}:
\bctitle{Adaptivepose: Human parts as adaptive points}.
In: \bbtitle{Proceedings of the AAAI Conference on Artificial Intelligence},
vol. \bseriesno{36},
pp. \bfpage{2813}--\blpage{2821}
(\byear{2022})
\end{bchapter}
\endbibitem

\bibitem[\protect\citeauthoryear{Geng et~al.}{2021}]{geng2021bottom}
\begin{bchapter}
\bauthor{\bsnm{Geng}, \binits{Z.}},
\bauthor{\bsnm{Sun}, \binits{K.}},
\bauthor{\bsnm{Xiao}, \binits{B.}},
\bauthor{\bsnm{Zhang}, \binits{Z.}},
\bauthor{\bsnm{Wang}, \binits{J.}}:
\bctitle{Bottom-up human pose estimation via disentangled keypoint regression}.
In: \bbtitle{Proceedings of the IEEE/CVF Conference on Computer Vision and Pattern Recognition},
pp. \bfpage{14676}--\blpage{14686}
(\byear{2021})
\end{bchapter}
\endbibitem

\end{thebibliography}

\end{document}